\documentclass[10pt]{article} 
\usepackage[accepted]{tmlr}

\usepackage[utf8]{inputenc} 
\usepackage[T1]{fontenc}    
\usepackage{hyperref}
\usepackage{url}            
\usepackage{booktabs}       
\usepackage{amsfonts}       
\usepackage{nicefrac}       
\usepackage{microtype}      
\usepackage[table]{xcolor}         
\usepackage{graphicx}%
\usepackage{multirow}%
\usepackage{amsmath,amssymb,amsfonts}%
\usepackage{amsthm}%
\usepackage{mathrsfs}%
\usepackage{enumitem}
\usepackage{textcomp}%
\usepackage{manyfoot}%
\usepackage{subcaption}
\usepackage{cleveref}
\usepackage{listings}
\usepackage[export]{adjustbox} 
\usepackage{csquotes}
\usepackage{algorithm}
\usepackage{algpseudocode}
\usepackage{numprint}
\usepackage{mathtools}
\usepackage{xspace}
\usepackage[acronym]{glossaries}
\usepackage{placeins} 
\usepackage[labelfont=bf,tableposition=top]{caption}
\DeclareCaptionLabelFormat{andtable}{#1~#2  \&  \tablename~\thetable}
\DeclareCaptionLabelFormat{andfigure}{\tablename~\thetable  \ \&  #1~#2}

\newcommand{\task}{\texttt{T}\xspace}
\newcommand{\model}{\texttt{FM}\xspace}
\newcommand{\GEN}{\texttt{GEN}\xspace}

\newcommand{\NORMs}{\texttt{NORM}\xspace}
\newcommand{\NORMpre}{\texttt{NORM}\textsubscript{\textit{pre}}\xspace}
\newcommand{\NORMpost}{\texttt{NORM}\textsubscript{\textit{post}}\xspace}
\newcommand{\NORMinf}{\texttt{NORM}\textsubscript{\textit{inf}}\xspace}
\newcommand{\AGG}{\texttt{AGG}\xspace}

\newcommand{\none}{\texttt{none}\xspace}
\newcommand{\all}{\texttt{all}\xspace}
\newcommand{\mean}{\texttt{mean}\xspace}
\newcommand{\krandom}{\texttt{k-random}\xspace}
\newcommand{\kcovmax}{\texttt{k-cov-max}\xspace}
\newcommand{\kmedoids}{\texttt{k-medoids}\xspace}
\newcommand{\kmeans}{\texttt{k-means}\xspace}
\newcommand{\ls}{\texttt{least-squares}\xspace}
\newcommand{\kls}{\texttt{k-least-squares}\xspace}
\newcommand{\kfps}{\texttt{k-fps}\xspace}

\newcommand{\fewshot}[1]{\texttt{#1-shot}}

\renewcommand{\ll}{\texttt{L2}\xspace}

\newcommand{\quant}{\texttt{quantile}\xspace}
\newcommand{\nn}[1]{\texttt{#1-nn}\xspace}
\newcommand{\aggmax}{\texttt{max}\xspace}

\newcommand{\resnetsmall}{\texttt{resnet18}\xspace}
\newcommand{\resnetlarge}{\texttt{resnet50}\xspace}
\newcommand{\vitb}{\texttt{vit\_b\_16}\xspace}
\newcommand{\swinb}{\texttt{swin\_b}\xspace}

\newcommand{\imagenet}{\textit{ImageNet}\xspace}
\newcommand{\mnist}{\textit{MNIST}\xspace}
\newcommand{\fmnist}{\textit{FashionMNIST}\xspace}
\newcommand{\cifar}{\textit{CIFAR-10}\xspace}
\newcommand{\textall}{\text{all}}
\newcommand{\mnistm}{\textit{MNIST-M}\xspace}
\newcommand{\svhn}{\textit{SVHN}\xspace}
\newcommand{\usps}{\textit{USPS}\xspace}
\newcommand{\combidigits}{\textit{CombiDigits}\xspace}

\newcommand{\framework}{\texttt{IMPRINT}\xspace}

\newcommand{\ncone}{$\mathcal{NC}_1$\xspace}

\newcommand{\vect}[1]{\boldsymbol{\mathbf{#1}}}

\newcommand{\R}{\mathbb{R}}
\newcommand{\N}{\mathbb{N}}

\title{Robust Weight Imprinting: Insights from Neural Collapse and Proxy-Based Aggregation}

\author{%
  Justus Westerhoff$^{1}$ \quad Golzar Atefi$^{1}$ \quad Mario Koddenbrock$^{2}$ \quad Alexei Figueroa$^{1}$ \\
  \textbf{Alexander L\"oser}$^{1}$ \quad \textbf{Erik Rodner}$^{2}$ \quad \textbf{Felix A.~Gers}$^{1}$ \\
  $^1$DATEXIS, Berliner Hochschule f\"ur Technik (BHT), Germany \\
  $^2$KI-Werkstatt, University of Applied Sciences Berlin, Merantix Momentum, Germany \\
  \texttt{\{justus.westerhoff,golzar.atefi\}@bht-berlin.de}
}

\author{\name Justus Westerhoff \email justus.westerhoff@bht-berlin.de \\
      \addr DATEXIS, Berliner Hochschule f\"ur Technik (BHT), Germany
      \AND
      \name Golzar Atefi \\
      \addr DATEXIS, Berliner Hochschule f\"ur Technik (BHT), Germany
      \AND
      \name Mario Koddenbrock \\
      \addr KI Werkstatt, Hochschule für Technik und Wirtschaft Berlin (HTW), Germany
      \AND
      \name Alexei Figueroa \\
      \addr DATEXIS, Berliner Hochschule f\"ur Technik (BHT), Germany
      \AND
      \name Alexander L\"oser \\
      \addr DATEXIS, Berliner Hochschule f\"ur Technik (BHT), Germany
      \AND
      \name Erik Rodner \\
      \addr KI Werkstatt, Hochschule für Technik und Wirtschaft Berlin (HTW) \& Merantix Momentum, Germany
      \AND
      \name Felix A.~Gers \\
      \addr DATEXIS, Berliner Hochschule f\"ur Technik (BHT), Germany
}

\def\month{12}  
\def\year{2025} 
\def\openreview{\url{https://openreview.net/forum?id=duU11BnQ3Y}} 

\begin{document}

\maketitle

\begin{abstract}
    The capacity of foundation models allows for their application to new, unseen tasks.
    The adaptation to such tasks is called transfer learning.
    An efficient transfer learning method that circumvents parameter optimization is imprinting.
    The conceptual differences between studies on imprinting form the basis of our systematic investigation.
    In this work, we propose the general \framework framework, identifying three main components: generation, normalization, and aggregation.
    Through the lens of this framework, we conduct an in-depth analysis and a comparison of the existing methods. 
    Our findings reveal the benefits of representing novel data with multiple proxies in the generation step and show the importance of proper normalization.
    Beyond an extensive analytical grounding, our framework enables us to propose a novel variant of imprinting which outperforms previous work on transfer learning tasks by 4\%. 
    This variant determines proxies through clustering motivated by the neural collapse phenomenon -- a connection that we draw for the first time.
    We publicly release our code at \url{https://github.com/DATEXIS/IMPRINT}.
\end{abstract}

\section{Introduction}
\label{sec:introduction}
In machine learning applications, training models from scratch is often not viable due to limitations in data and compute.
A popular solution is to apply transfer learning \citep{transferlearning3,transferlearning2} based on foundation models (\model{s}) \citep{foundationmodels} that are pre-trained on a large amount of data.
To tune a \model to a novel task, e.g., for classification, a common procedure is to freeze the model parameters and replace the output layer with a new head.
A particularly simple method for implementing such a new head was proposed by \citet{qi-imprinting} and coined \textit{imprinting}.

\paragraph{Imprinting.}
In the original work by \citet{qi-imprinting},
the last-layer weight vector of a novel class is set to the normalized average of its scaled embedding vectors, i.e., its class mean.
These class means are representatives of the classes, which we denote as \textit{proxies}.
In general, we refer to imprinting as efficient learning methods without the need for cross-class statistics or gradient-based optimization.
A plethora of studies have emerged surveying this technique by adding complexity and adaptability \citep{fewshot-metagenweights,masked-imprinting,hypersphere-imprinting, imprint-multilabel, imprint-objdetec, multiple-projection-head-imprinting, yan2023few}.
Despite its many adaptations, imprinting lacks a systematic comparison that unifies them.
Understanding its variations could unlock greater efficiency across many fields, making the method even more versatile and impactful.

\paragraph{Applications.}
In particular, while in some practical applications, accuracy is prioritized over computational efficiency, the latter does become a critical requirement in scenarios where computational resources are severely constrained, e.g., in the chemical and polymer processing industries.
Here, battery-powered edge compute is essential and frequent retraining or large-scale optimization is infeasible. 
Imprinting has proven particularly effective in these contexts of edge-embedded devices \citep{hypersphere-imprinting}.
For instance, \citet{imprintingrobot} implement a vision-based robotic force grasping with a variable-stiffness gripper that can safely handle both fragile and heavy objects by rapidly adapting to novel categories without retraining (Continual Learning (CL) setting).
Industrial adoption also underscores this trend with Google’s Coral Edge TPU including an \texttt{ImprintingEngine} API~\citep{coralai}, that allows users to add new classes from a few examples without recompiling the model.
Very recently, \citep{belal2025fsid} apply imprinting to spectrogram embeddings from IMU/EMG gait data, achieving efficient classification in low-data Human Activity Recognition (HAR) tasks.
Apart from that, \cite{ncmbaseline} extend imprinting to CL, establishing a simple baseline with competitive performance compared to more sophisticated state-of-the-art CL algorithms.

\begin{figure}[t]
    \vspace*{-5mm}
    \includegraphics[width=0.77\linewidth,center,trim=20 15 22 15,clip]{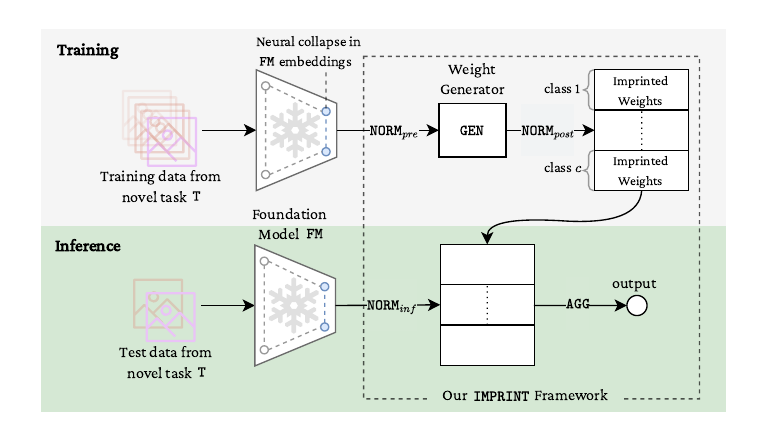} 
    \caption{
        Overview of our \framework framework. 
        The foundation model \model is frozen and shows neural collapse.
        The weight generator (\GEN) uses training data from a novel task \task to consecutively generate one or more weight vectors (proxies) per class $1, \dotsc, C$ in \task.
        In inference, the final output for the test data in \task is computed by an aggregation (\AGG) mechanism.
        Embeddings and generated weights are normalized according to \NORMpre and \NORMpost, respectively.
        During inference, embeddings are normalized according to \NORMinf.
    }
    \label{fig:framework}
\end{figure}

\begin{figure}[t]
    \captionlistentry[table]{A table beside a figure}
    \captionsetup{labelformat=andfigure}
    \caption{
    Previously studied imprinting strategies are special cases within \framework.
    We evaluate 12 different classification tasks \task{s} derived from \mnist, \fmnist, and \cifar, each with 10 classes or subsets thereof, and 4 pre-trained models \model{s} (\resnetsmall, \resnetlarge, \vitb, \swinb).
    The proposed configuration (\enquote{Ours}) derived from \framework outperforms previous work across \model{s} and \task{s} by a large margin with statistical significance, as confirmed by the critical difference (CD) diagram below.
    Since absolute accuracies vary substantially across tasks and models (as reflected by the large standard deviations (std)), this rank-based aggregation is used for fair comparison.
    Here, $k=20$ is used, highlighting the gain of using multiple proxies per class.
    For reference, the gray row reports an oracle method that uses cross-class feature statistics to generate weights (see \cref{ss:optimalweights}).
    It is not an imprinting method and therefore not directly comparable to the imprinting-based approaches above.
    Nonetheless, the results indicate that our method substantially narrows the gap between single-proxy \mean imprinting and this oracle baseline.
    }
    {
    \begin{tabular}{llllllr}
            Work & \NORMpre & \GEN & \NORMpost & \NORMinf & \AGG & Avg. acc. \% {\scriptsize $\pm$ std} \\
            \midrule
            {\citet{qi-imprinting}}   & \ll     & \mean  & \ll     & \ll     & \aggmax & 86.79 {\scriptsize $\pm 7.83$} \\
            {\citet{itsdone}} & \none   & \mean  & \quant  & \none   & \aggmax & 82.90 {\scriptsize $\pm 12.87$} \\
            \citet{ncmbaseline} & \none   & \mean  & \none   & \none   & \nn{1} & 86.64 {\scriptsize $\pm 7.88$} \\
            Ours & \ll & \kmeans & \ll & \ll & \aggmax & \textbf{91.06} {\scriptsize $\pm 6.21$} \\
            \midrule
            \rowcolor{black!4}
            Oracle & \none & \ls & \none & \none & \aggmax & 94.54 {\scriptsize $\pm 5.00$} \\
            \midrule
    \end{tabular}
    }
    \centering
    \includegraphics[width=0.6\linewidth]{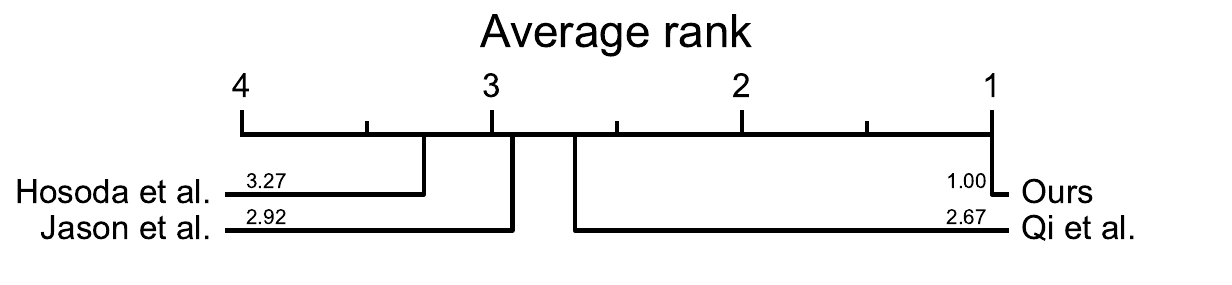}
    \label{tab:special-cases}
\end{figure}

\paragraph{Framework.}
We present \framework, a framework that enables a comprehensive analysis of existing imprinting techniques.
More precisely, we generalize prior work by decomposing imprinting into three principal steps (see \cref{fig:framework}).
During generation (\GEN) of weights, the method selects representative data samples and generates one or more weight vectors (proxies) per class.
Normalization (\NORMs) is crucial, as the generated weights need to be balanced.
Aggregation (\AGG) entails the computation of the final output, e.g., a class label.
The computational efficiency of imprinting allows us to perform a large number of experiments.
Through \framework, we are able to propose a novel, best-performing imprinting strategy using multi-proxy weight imprinting in combination with $L^2$ normalization, outperforming previously studied methods, as depicted in \cref{tab:special-cases}.

\paragraph{Neural Collapse.}
When neural networks are trained to reach near-zero loss, their penultimate-layer embeddings collapse to the class means \citep{papyan2020prevalence, zhu2021geometric}.
We investigate this phenomenon as a potential explanation for when and why imprinting works.
Our analysis proves that there exists a relationship between a measure of neural collapse and the success of imprinting.
Since quantification of this phenomenon is possible in a post-hoc fashion over the \model features, we believe that these insights will contribute to further development of imprinting methods, as well as, training regimes of \model{s} that are more suitable for transfer learning via imprinting.

\paragraph{Contributions.} In summary, our main results and contributions are:
\begin{itemize}[itemsep=2pt, parsep=0pt]
    \vspace*{-2mm}
    \item We deconstruct weight imprinting into the \framework framework composed of generation, normalization, and aggregation, and discuss variations for each of them, while identifying prior work as special cases (\cref{sec:method}).
    To the best of our knowledge, we are the first to conduct a comprehensive analysis of imprinting to this scale (\cref{sec:experiments}).
    \item We present a new, outperforming imprinting method utilizing $k$-means clustering for weight generation (\cref{ss:framework-results}) and show its benefits in certain low-data regimes (\cref{ss:lowdata}).
    \item As far as we are aware, we are the first to identify a relationship between the degree of neural collapse and imprinting success (\cref{ss:nc-and-acc}).
\end{itemize}

\FloatBarrier

\section{Related Work}
\label{sec:related}

\paragraph{Imprinting and Low-Data Regimes.} 
Weight imprinting is implemented by setting the final layer weights for the novel classes to the scaled average of the embedding vectors of their training samples and was first introduced by \citet{qi-imprinting} for the few-shot learning scenario.
The authors find that for up to $20$ samples, using a combination of imprinting and fine-tuning outperforms other state-of-the-art methods, including nearest neighbor algorithms.
In contract, in our work we do not limit the number of samples and perform no fine-tuning on the imprinted weights to maintain efficiency.

\par
Imprinting has been applied to object detection \citep{imprint-objdetec, yan2023few}, multi-label classification \citep{imprint-multilabel}, semantic segmentation \citep{masked-imprinting}, and combined with an attention mechanism to generate weights for the novel classes in a few-shot classification task \citep{fewshot-metagenweights}.

\par
\citet{itsdone} apply imprinting using quantile normalization to ensure statistical similarity between new and existing weights.
We consider this as one normalization scheme in our framework.
\citet{covidimprint} apply imprinting in chest radiography for detection of COVID-19 and find that it yields better results than joint gradient descent training of all classes when only few samples are available.
They speculate whether normalization is a constraint in their imprinting model.

\par
Before the era of deep learning, \citet{first-nearestclassmean} analyze the transferability of hand-crafted image features.
They use a \enquote{nearest class multiple centroids} (NCMC) classifier with multiple proxies generated from a $k$-means clustering algorithm.
In combination with metric learning, they compare favorably against the $m$-nearest neighbor algorithm.
Our work, on the other hand, highlights efficient transfer learning provided by foundation models.

\paragraph{Transfer Learning.}
Using embedding vectors from pre-trained models is a simple and widely used transfer learning approach, established in the seminal works on computer vision~\citep{donahue_decaf_2014} and natural language processing~\citep{devlin_bert_2019}. 
\citet{kornblith_better_2019} show that pre-training performance of a model is highly correlated with the performance of the resulting embedding vectors in downstream tasks.
In addition, \citet{huh_what_2016} provide insights into the required quality of pre-training data.
Our work is orthogonal to these studies, since we focus on studying weight generation, normalization, and aggregation techniques applied later on for new task adaptation. 

\paragraph{Continual Learning (CL).}
Although we investigate transfer learning scenarios, we review the imprinting applications and results from CL. 
\citet{icarl} dynamically select a subset of examples for each class and update internal representations via gradient descent.
They use a nearest mean classifier (NMC) with respect to the saved examples.
\citet{ncmbaseline} use an NMC classifier as well and achieve good performance on CL benchmarks without any fine-tuning of the embeddings.
However, they do not investigate the effect of normalization and using multiple proxies.

\par
Findings of \citet{imprintallknn} show that a simple, approximate $m$-nearest neighbor classifier outperforms existing methods in an Online CL setting when all data can be stored.
In our work, however, we compare imprinting all data to a limited number of more representative proxies striving for efficiency. 

\paragraph{Neural Collapse (NC).} The phenomenon of NC was identified by \citet{papyan2020prevalence} and refers to the convergence of the last-layer weight vectors to class means.
It was shown that, regardless of the loss function, optimizer, batch-normalization, or regularization, NC will eventually occur (provided the training data has a balanced distribution) \citep{zhu2021geometric, neuralcollapse2, neuralcollapse3}.
Nevertheless, complete neural collapse is practically unrealistic \citep{DBLP:conf/icml/TirerHN23}.
In transfer learning, \citet{DBLP:conf/iclr/Galanti0H22} show that NC occurs on new samples and classes from the same distribution as the pre-training dataset, highlighting the usability of foundational models in such scenarios.
In our work, we expand the survey on NC by experimenting with out-of-distribution classes belonging to different datasets and linking their degree of collapse to the success of certain imprinting strategies.


\section{Methods}

\subsection{\texttt{IMPRINT}}
\label{sec:method}
In order to find out how to best and efficiently set the classifier weights of a foundational model \model in new, previously unseen tasks \task, we create the \framework framework (see \cref{fig:framework}) that generalizes previous work, specifically all the methods that work without access to cross-class statistics and gradient-based training.
Thereby, we unify all the existing imprinting strategies described in \cref{sec:related}.

\paragraph{Overview.}
We analyze the effect of weight generation (\GEN), normalizations ($\NORMs = \{\NORMpre, \NORMpost, \NORMinf\}$), and aggregation (\AGG).
The \framework framework depicted in \cref{fig:framework} consists of three main building blocks: a foundation model \model, a weight generator \GEN, and extendable classifier weights that are imprinted.
The \model remains frozen throughout the experiments.
It receives data from \task as inputs and produces embedding vectors.
The training process generates weight vectors for each of the $C$ classes in \task.
Hereby, embeddings from the \model are normalized before the generation (\GEN) step according to \NORMpre.
The generated weight vectors per class are referred to as \textit{proxies}, prototypes, or representatives \citep{metric-proxies, prototypical, convolutional-prototypes, semanticdriftCL}.
These proxies are normalized according to \NORMpost.
As in the work by \citet{qi-imprinting}, we do not use bias values.
To classify the test data in \task during inference, it is first embedded by the \model, normalized according to \NORMinf, and finally aggregated by \AGG, resulting in a predicted class label.

\paragraph{Generalization of Previous Methods.}
Previously proposed imprinting methods can be defined as special cases of our framework.
We design \framework such that every method can be defined by a single combination of \GEN, \NORMs, and \AGG, of which we inspect all possible combinations. 

\paragraph{Weight Generation (\GEN).}
The purpose of \GEN is to determine how the embeddings of the training data in \task are used to form the new weights.
In contrast to \citet{qi-imprinting} which only incorporates one proxy per class (the mean), we add flexibility by allowing each class to have multiple proxies as in \citet{first-nearestclassmean} and enable non-linear classification.
We denote the number of proxies per class as $k$, ranging between $1$ and the number of samples, and investigate the following operations \textbf{conducted per class} to generate those:

\begin{itemize}[itemsep=2pt, parsep=0pt]
    \item \all: All embeddings (denoted as $k=\textall$).
    \item \krandom: $k$ random embeddings.
    \item \mean: The mean of all embeddings.
    \item \kmeans: $k$-means cluster centers using \texttt{KMeans} from \textit{sklearn}~\citep{scikit-learn}. $k=1$ is the same as \mean.
    \item \kmedoids: $k$-medoids cluster centers using \texttt{KMedoids} from \textit{sklearn}.
    \item \kcovmax (covariance-maximization): Top $k$ embeddings by covariance (in code: \lstinline[language=Python]|proxies = embeddings[torch.argsort(torch.sum(torch.cov(embeddings), dim=0), descending=True)[:k]]|).
    \item \kfps (farthest-point sampling): Iteratively selecting $k$ embeddings, such that it maximizes the distance from already selected ones (starting with a random sample).
\end{itemize}
We choose this diverse list of methods to cover a wide range of approaches, ranging from heuristics (e.g., \kfps) to more complex algorithms (e.g., \kmeans).
Note that only \mean and \kmeans generate proxies beyond existing samples by producing synthetic cluster centers in embedding space.
In comparison, \kmedoids must choose actual samples as cluster centers (analogous to a median).
None of these methods use cross-class statistics or gradient-based optimization.
We also note an analogy to associative memory models, interpreting imprinting as a memory update process following a covariance rule, as detailed in \cref{ss:imprinting-as-memory}.

\paragraph{Normalization (\NORMs).} The main reason for applying normalization is to allow each embedding and weight vector to contribute equally on the same scale.
The modes we allow are no normalization (\none), $L^2$ normalization (\ll), and quantile normalization (\quant).

\ll normalization can be applied to embeddings before \GEN via \NORMpre, to the generated weights via \NORMpost, and to embeddings in inference via \NORMinf. In any case, the vector is $L^2$-normalized by dividing it by its Euclidean length $\Vert \cdot \Vert_2$.

\quant normalization \citep{quantnorm1,quantnorm2} can only be applied to generated weights.
This non-linear operation distributes weights equally.
Recall that if more than one class is contained in \task ($c > 1$), \GEN is performed for each class consecutively, and the reference distribution changes accordingly.
In particular, for the first class there is no reference distribution to map to.
This is different from \citet{itsdone}, where new weights are matched to the distribution of the original classifier weights of the \model.
Since we do not consider the classes used for pre-training the \model and especially do not assume access to their last-layer weights, this is not possible in our scenario.

\paragraph{Aggregation (\AGG).} There are various ways to use the generated weights (proxies) per class during inference, especially when $k>1$.
We focus on two different modes, \aggmax and $m$-nearest neighbor (\nn{m}).
The former, \aggmax, computes the inner product of the input embedding and the imprinted weights and outputs the class label with the maximum activation.
The latter, \nn{m}, uses the class weights as keys and the embeddings as values, and chooses the final winning output class via the $m$-nearest neighbor algorithm.
The \nn{m} voting is weighted by the inverse of the Euclidean distances to their nearest neighbor, turning it into weighted majority voting.

We also experimented with alternative distance functions (Cosine, Manhattan) and uniform weighting, but found differences well within statistical noise.
Using the Mahalanobis distance proved computationally prohibitive. 
More elaborate voting mechanisms, such as learned top-$k$ attention or entropy-based filtering, are conceivable, but extend well beyond the minimalist imprinting paradigm and therefore left for future work.

\par
Note that \aggmax is the same as \nn{1} in the case of \ll for \NORMpost, since for any fixed embedding vector $\vect{v}$ and variable proxy $\vect{w}$, the argmin of
$\Vert \vect{v} - \vect{w} \Vert^2 = \Vert \vect{v} \Vert^2 - 2\langle \vect{v}, \vect{w} \rangle + \Vert \vect{w} \Vert^2$,
calculated by \nn{1}, is the same as the argmax of the inner product $\langle \vect{v}, \vect{w}\rangle$ calculated in \aggmax.

\subsection{Quantifying Neural Collapse}\label{ss:quantnc}
\begin{figure}[htbp]
    \includegraphics[page=1,width=1.08\linewidth,center]{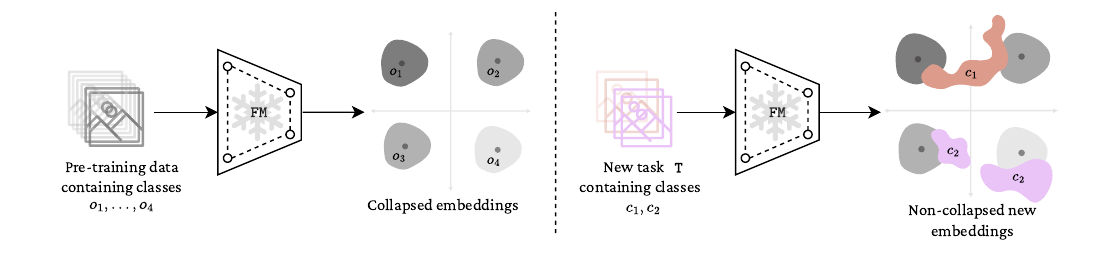}
    \caption{Left: The embeddings of the pre-training data, after being used to train the foundation model \model, show neural collapse, as each class ($o_1,\dotsc,o_4$) is evenly separated in space and accumulates around their respective class means. Right: For a novel task with classes $c_1,c_2$ (pink and brown) scatter around the collapsed pre-trained classes (gray). 
    }
    \label{fig:nc}
\end{figure}

Neural collapse (NC)~\citep{papyan2020prevalence} characterizes the state of the features produced by a classification neural network  after training to near zero training loss.
Namely, the learned embeddings of each class converge, i.e., \textit{collapse}, to their class means.
These globally centered class means and classifier weights form a simplex equiangular tight frame (ETF) -- a collection of equal length and maximally equiangular vectors, that maximize the between-class variability.
This results in an optimal linearly separable state for classification.
In \cref{fig:nc} (left), we illustrate the collapse of a \model on its pre-training data.
The newly arrived data \task from a different dataset is distributed more unevenly across the embedding space (right).

\par
Two important characteristics of NC are \textbf{variability collapse}, i.e., the within-class variability of the penultimate-layer embeddings collapses to zero, and \textbf{convergence to nearest-mean-classification}.
We focus on variability collapse (\ncone) as in \citet{zhu2021geometric}.
Given a foundation model \model{} and a finite dataset containing $C$ classes with (for simplicity) $N$ samples per class, we use its $l$-dimensional embeddings $\{\vect{h}_{c,i}\}_{1\le c \le C,\; 1 \le i \le N}$ given by \model{} to define the global feature mean $\vect{h}_G \coloneqq \frac{1}{CN}\sum_{c=1}^{C}\sum_{i=1}^{N} \vect{h}_{c,i}$, class means $\overline{\vect{h}}_c \coloneqq \frac{1}{N}\sum_{i=1}^{N} \vect{h}_{c,i}$, the within-class covariance matrix $\mathbf{\Sigma}_{W} \coloneqq \frac{1}{CN}\sum_{c=1}^{C}\sum_{i=1}^{N}\bigl(\vect{h}_{c,i}-\overline{\vect{h}}_c\bigr)\bigl(\vect{h}_{c,i}-\overline{\vect{h}}_c\bigr)^{\top}$, and the between-class covariance matrix $\mathbf{\Sigma}_{B} \coloneqq \frac{1}{C}\sum_{c=1}^{C}\bigl(\overline{\vect{h}}_c-\vect{h}_{G}\bigr)\bigl(\overline{\vect{h}}_c-\vect{h}_{G}\bigr)^{\top}$ to finally compute
\begin{equation}\label{eq:ncone}
\mathcal{NC}_{1} = \tfrac{1}{C} \mathrm{trace}(\mathbf{\Sigma_{W}} {\mathbf{\Sigma}^{+}_{B}}),
\end{equation}
where $^{+}$ symbolizes the pseudo-inverse.

Based on \cref{eq:ncone}, an \ncone score closer to zero signifies a higher collapse.
In contrast, an increase in multi-modality of data leads to a higher \ncone score (as analyzed in \cref{fig:nc1s}).
Note that this measurement is not independent of the embedding dimension $l$ and the number of classes $C$.
According to NC, imprinting the mean, as originally done by \citet{qi-imprinting}, is best when \ncone is small.
We claim that when the data is not fully collapsed (as is often the case in practice), the scale of \ncone could guide the proxy generation method, e.g., having multiple proxies $k>1$ per class.
We investigate this in \cref{ss:nc-and-acc}.

\subsection{Significance Testing with Critical Difference Diagrams}\label{subsec:cd-diag}

Our experiments compare large numbers of imprinting configurations across tasks with substantially different achievable accuracy levels.
To obtain scale-independent statistical comparisons, we base our analysis on ranking (dis-)agreements rather than raw accuracy.
Non-parametric rank-based tests are well suited for this setting.
Following \citet{cddiag}, we use the Friedman test and Wilcoxon signed-rank tests because they do not assume normality or homoscedasticity, and can be applied uniformly to any evaluation metric. Empirical evidence in \citet{cddiag} further suggests that these tests have good power in classifier comparison scenarios.

Concretely, for each evaluation instance (i.e., foundation model \model and task \task) we compute the classification accuracy of all configurations and assign ranks (rank 1 = highest accuracy).
This yields a matrix of ranks with one row per configuration and one column per evaluation instance.
On this matrix we first apply the Friedman test to assess whether there are any overall differences between configurations.
If the Friedman test rejects the global null hypothesis at significance level $\alpha = 0.05$, we proceed with a post-hoc, pairwise analysis.
Specifically, for every pair of configurations we run a two-sided Wilcoxon signed-rank test across evaluation instances.
The resulting $p$-values are corrected for multiple comparisons using the Holm step-down procedure~\citep{holm1979simple}. 

Critical difference (CD) diagrams summarize both the average performance and the significance structure.
The horizontal axis shows the average rank of each configuration across all evaluation instances, with lower ranks indicating better performance.
Two configurations are connected by a thick horizontal line if their Wilcoxon--Holm comparison does not show a significant difference.

\section{Experimental Setup}
\label{sec:experiments}

\paragraph{Foundation Models \model.}
We use \resnetsmall, \resnetlarge~\citep{resnet}, \vitb~\citep{vit}, and \swinb~\citep{swin} as \model{s}, two CNN-based and two Transformer-based architectures.
All four models are pre-trained on \imagenet{-1K} (ILSVRC 2012) \citep{imagenet}.
To generate the embeddings, we use PyTorch's \textit{torchvision} models.

\paragraph{Tasks \task.}
We analyze multi-class classification scenarios without separating base and new classes, instead focusing on all classes within a novel \task at the same time.
To investigate the effect of the number of samples given, we look at \fewshot{n} ($n \in \N$) scenarios.
For that, we randomly pre-sample the training data of \task to $n$ samples per class -- transitioning into the low-data regime.

\par
To find out the best imprinting strategy within our \framework framework, we focus on tasks \task created from the datasets \mnist~\citep{mnist}, \fmnist~\citep{fashionmnist}, and \cifar~\citep{cifar10}, each containing 10 classes.
We mainly focus on the three \task containing all ten classes.
Furthermore, we look at smaller tasks only containing classes $\{0,1,2\}$, and the two tasks containing classes $\{1,3,5,7,9\}$ resp. $\{0,2,4,6,8\}$.
This random selection of $3 \cdot 4 = 12$ tasks adds variation to our evaluations.

\begin{figure}[htbp]
    \includegraphics[width=0.8\linewidth,center,trim=0 10 0 10,clip]{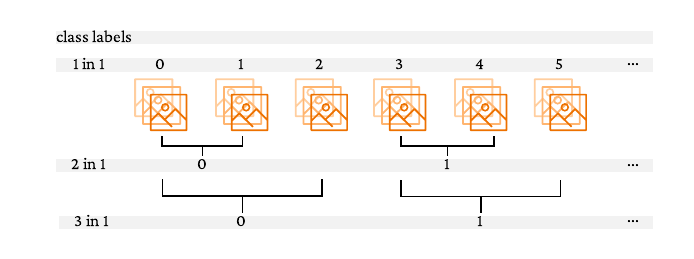} 
    \caption{Combining multiple classes into one to create tasks with multi-modal class distributions. Simplified example for \enquote{$d$ in 1}, $d=1,2,3$, with only six (instead of 100) original classes.}
    \label{fig:howto-label-remapping}
\end{figure}

\paragraph{Neural Collapse.}
In the analysis of neural collapse (NC), we also look at the \model{s}' pre-training data (\imagenet).
As its test set is not available, we use its validation set in \ncone computations.
Furthermore, for \imagenet, we relabel data by combining multiple classes into one label to simulate multi-modal class distributions for an in-depth NC analysis.
These tasks are called \enquote{$d$ in $1$}, $d=1,\dotsc,10$, each containing $10$ different labels.
More precisely, we take 100 random classes from \imagenet and sequentially map the first $d$ to label $1$, the second $d$ to label $2$, etc., until we reach 10 distinct labels.
See \cref{fig:howto-label-remapping} for a simplified illustration.

\par
Moreover, we construct a new out-of-distribution, non-collapsed dataset by merging all classes from four digit datasets: \mnist, \mnistm~\citep{mnistm}, \svhn~\citep{svhn}, and \usps~\citep{usps}.
The resulting dataset, which we refer to as \combidigits, exhibits reduced collapse due to the greater distributional diversity within each class.
To ensure scale invariance in covariance-based NC measurements, all embeddings are $L^2$-normalized before computing \ncone.

\paragraph{Scale.}
In total, we ran approximately \numprint{500000} experiments, varying imprinting components, foundation models, tasks, and seeds.
This is feasible with minimal effort as imprinting is a highly efficient method: core steps such as weight generation (\GEN), normalization (\NORMs), and aggregation (\AGG) are linear in dataset size or number of proxies and parallelize efficiently within each step.
Details on experimental infrastructure and parallelization are provided in \cref{ss:comp_efficiency}.

\paragraph{Evaluation.}
Unless stated otherwise, we report the median test accuracy over three different seeds.
In \cref{ss:framework-results,ss:lowdata}, we evaluate the imprinting performance by varying the \model (4) and \task (12) and report average accuracy and standard deviation (std) across these.
Due to the heterogeneity of models and tasks, large std are expected.
Therefore, overall method comparisons are based on ranks rather than absolute accuracies.
Methods are ranked by their median accuracy, yielding $4 \cdot 12 = 48$ potentially different ranks.
We report the average rank and assess statistical significance of ranking (dis-)agreements using critical difference (CD) diagrams as explained in \cref{subsec:cd-diag}.
The code used to generate these diagrams is inspired by \citet{cddiag-code}.

\par
In experiments with neural collapse (\cref{ss:nc-and-acc}), we investigate the four \model{s} on 100 random \imagenet tasks with remapped labels (as explained above) and the four tasks containing all of \mnist, \fmnist, \cifar, and \combidigits, respectively.


\section{Results}
\label{sec:results}
Our main experimental insights are:
\textbf{1.} Our \framework framework generalizes previous methods, and we find a new superior imprinting strategy (\cref{ss:framework-results}).
\textbf{2.} We show that our novel strategy is even beneficial in low-data regimes with as little as 50 samples per class (\cref{ss:lowdata}).
\textbf{3.} We identify a correlation between imprinting success utilizing multiple proxies and a measure of neural collapse (\cref{ss:nc-and-acc}).

\subsection{Best Imprinting Strategy}
\label{ss:framework-results}

We provide a comparison between existing memory-constrained methods used for imprinting on foundation models in \cref{tab:special-cases}, namely, the work by \citet{qi-imprinting,itsdone,ncmbaseline}, as well as a novel, best-performing configuration (\enquote{Ours}) that results from \framework.
Focusing on $k=20$, we find that our method, consisting of \kmeans weight generation, \ll normalizations, and \aggmax aggregation, outperforms all previously studied approaches by a margin of $4\%$ on average with statistical significance.
For reference, we additionally report an oracle baseline that uses cross-class feature statistics to generate its weights, representing an upper bound that is not constrained by imprinting.
The results indicate that our new imprinting method significantly narrows the gap between single-proxy \mean imprinting and this oracle baseline.

The impact of using the unconstrained \nn{m} aggregation on \all data is investigated at the end of this section.
Next, we analyze each of the components of \framework separately.

\par
\begin{figure}[htbp]
    \captionlistentry[table]{A table beside a figure}
    \captionsetup{labelformat=andfigure}
    \caption{Benchmarking \GEN mechanism for $k \le 20$ across \model{s} and \task{s}. Best \NORMs combination for each row used implicitly. \AGG is fixed to \aggmax. CD diagram proves that \kmeans weight generation is significantly better than all other methods.}
    {
    \begin{tabular}{llr}
            \GEN & $k$ & Avg. acc. \% {\scriptsize $\pm$ std} \\
            \midrule
            \kmeans & 20 & \textbf{91.04} {\scriptsize $\pm 6.26$} \\ 
            \kmedoids & 20 & 87.01 {\scriptsize $\pm 8.23$} \\
            \mean & 1 & 86.84 {\scriptsize $\pm 7.80$} \\
            \kcovmax & 20 & 83.98 {\scriptsize $\pm 9.56$} \\
            \krandom & 20 & 82.14 {\scriptsize $\pm 10.15$} \\
            \kfps & 20 & 65.56 {\scriptsize $\pm 12.32$} \\ 
    \end{tabular}
    }
    \centering
    \raisebox{-2em}{ 
        \includegraphics[width=0.6\linewidth]{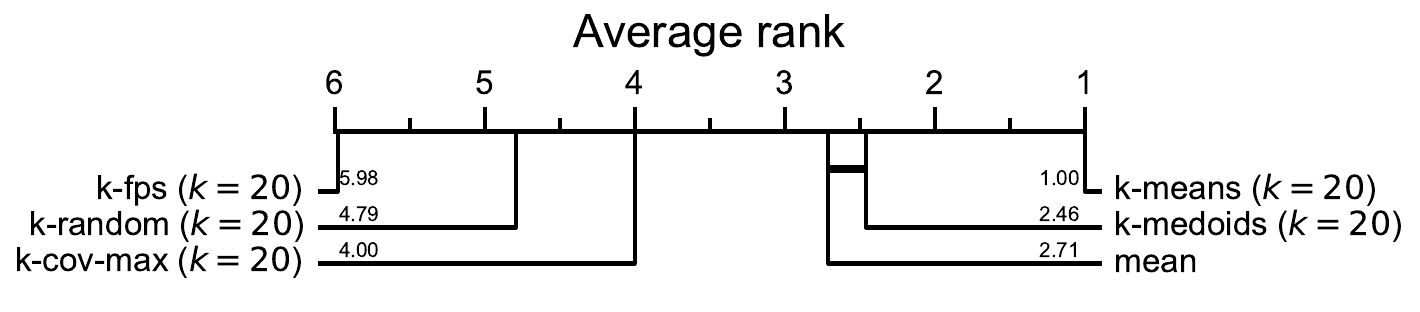}
    }
    \label{cd:proxyanalysis}
\end{figure}

\paragraph{Weight Generation (\GEN).}
To assess the impact of \GEN, we first focus on the \aggmax aggregation and do not fix \NORMs, but simply show the run with the best \NORMs combination, if not otherwise specified.
The $\nn{m}$ aggregation and different values for \NORMs are analyzed later in this section.

\par
Initially, we limit the number of generated proxies to $k \le 20$.
Results in \cref{cd:proxyanalysis} show how $\kmeans$, using as many proxies as possible (in this case, $20$) outperforms by $4\%$ on average accuracy compared to all the other \GEN methods.
The CD diagram illustrates its statistical significance in ranking.
Furthermore, while $\kmedoids$ with $20$ proxies (which necessarily have to be among the given samples, see \cref{sec:method}) is computationally expensive, it is on par with \mean, and covariance maximization, furthest-point sampling and random selection show even weaker performances.
We find similar results for $k \le 5$, where $\kmeans$ outperforms the other methods as well (see \cref{cd:proxyanalysis-less}).

\begin{figure}[tbhp]
    \centering
    \includegraphics[width=0.6\linewidth]{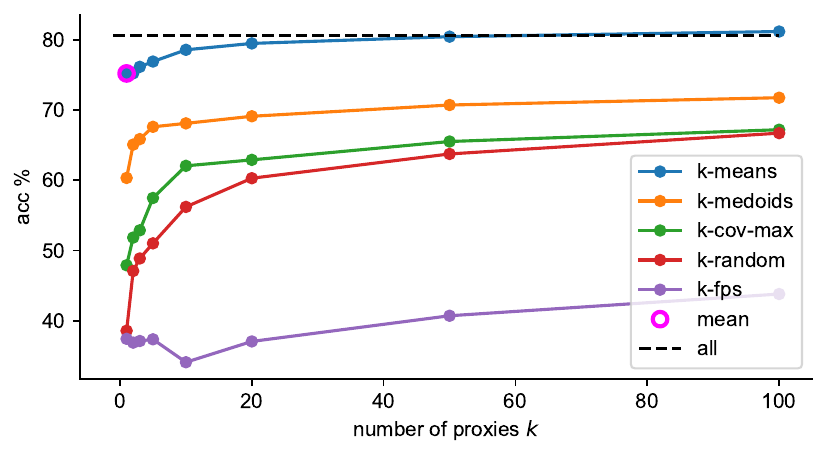}
    \caption{Benchmarking different \GEN methods with \resnetsmall on all of \cifar shows superiority of \kmeans proxies.
    All combinations employ \ll for all \NORMs and \aggmax as \AGG.}
    \label{fig:maxagg}
\end{figure}

\par
As the number of proxies $k$ increases, \kmeans continues to be the best \GEN method.
An example for \resnetsmall and \cifar can be found in \cref{fig:maxagg}.
All methods converge towards the point of imprinting (saving) \all data ($k=\textall$), even surpassing it in the case of \kmeans.
Due to its superior performance, we mainly focus on \kmeans in the remainder of the analysis.

\paragraph{Learned Weights.}
Employing gradient-based methods -- such as setting weights via the analytical least-squares initialization proposed by \citet{harun2025good}, which uses data \textit{and labels} from all classes \textit{jointly} -- can be seen as an upper bound (\enquote{Oracle} in \cref{tab:special-cases}) for the unsupervised weight imprinting approaches discussed here.
We extend this method by combining it with \kmeans, forming \kls, and observe in \cref{ss:optimalweights} that using multiple proxies per class instead of a single weight vector can still improve performance, particularly on datasets with high \ncone.

\begin{figure}[tbhp]
    \captionlistentry[table]{A table beside a figure}
    \captionsetup{labelformat=andfigure}
    \caption{Benchmarking \NORMpost mechanism across \model{s} and \task{s}. The best \NORMpre and \NORMinf combinations for each row are used implicitly. \GEN is fixed to \mean (that is, $k=1$) and \AGG is fixed to \aggmax. The CD diagram shows the statistical significance of \ll as the best weight normalization \NORMpost.}
    {
    \begin{tabular}{lr}
        \NORMpost & Avg. acc. \% {\scriptsize $\pm$ std} \\
        \midrule
        \ll & \textbf{86.84} {\scriptsize $\pm 7.80$} \\
        \quant & 82.90 {\scriptsize $\pm 12.87$} \\
        \none & 83.26 {\scriptsize $\pm 9.18$} \\
    \end{tabular}
    }
    \centering
    \raisebox{-3em}{ 
        \includegraphics[width=0.6\linewidth]{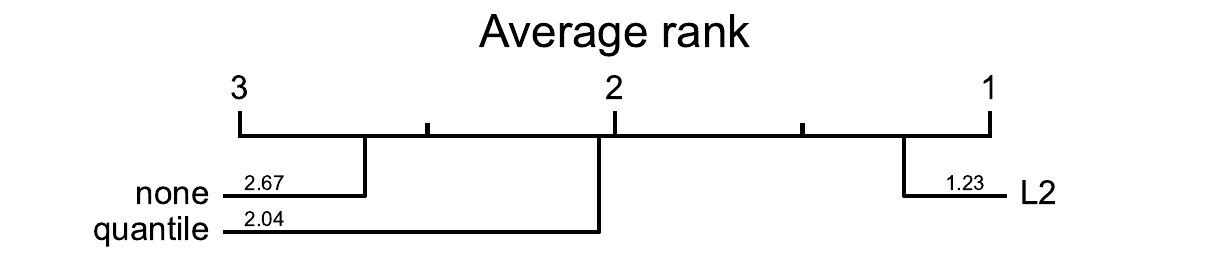}
    }
    \label{cd:normk1}
\end{figure}

\paragraph{Normalization (\NORMs).}
We compare all the different \NORMs methods, focusing on $\kmeans$ as \GEN.
For $k=1$ and varying \NORMpost (while taking best values for \NORMpre and \NORMinf implicitly), \cref{cd:normk1} shows that \ll is the best choice for weight normalization.
\quant and \none normalization both perform worse.

\par
\begin{table}[tbhp]
    \centering
    \captionsetup{width=\textwidth} 
    \caption{Benchmarking \NORMpre and \NORMinf mechanisms across \model{s} and \task{s}. \NORMpost is fixed to \ll, \GEN to \mean, and \AGG to \aggmax. No statistically significant differences were found.}
    {
    \begin{tabular}{llr}
        \NORMpre & \NORMinf & Avg. acc. \% {\scriptsize $\pm$ std} \\
        \midrule
        \none & \ll & 86.84 {\scriptsize $\pm 7.80$} \\
        \none & \none & 86.84 {\scriptsize $\pm 7.80$} \\
        \ll & \ll & 86.79 {\scriptsize $\pm 7.83$} \\
    \end{tabular}
    \label{tab:normk1-weightl2}
    }
\end{table}

Keeping \ll for \NORMpost fixed, we find no statistical differences between the different combinations of \NORMpre and \NORMinf (see \cref{tab:normk1-weightl2}).
For larger values of $k$, the differences among \NORMpost become even more pronounced.
However, the performance of \NORMpre and \NORMinf remains statistically indifferent for \ll weight normalization (see \cref{cd:normk1-full} for all combinations at once with $k=1$, and \cref{cd:normk20} for $k=20$).

\par
We restrict all the subsequent experiments to using \ll across all \NORMs following \citet{qi-imprinting}.
This combination of normalizations is chosen to specifically model cosine similarity within \aggmax aggregation.

\begin{figure}[tbhp]
    \captionlistentry[table]{A table beside a figure}
    \captionsetup{labelformat=andfigure}
    \caption{Benchmarking \AGG mechanism across \model{s} and \task{s}. \GEN is fixed to \all ($k=\textall$), that is, imprinting (saving) all data to weights. \ll normalization is used for all \NORMs. The CD diagram shows statistical significance of \nn{3}, \nn{5}, and \nn{20} over \aggmax aggregation.}
    {
    \begin{tabular}{lr}
        \AGG & Avg. acc. \% {\scriptsize $\pm$ std} \\
        \midrule
        \nn{5} & \textbf{93.74} {\scriptsize $\pm 4.97$} \\
        \nn{3} & 93.50 {\scriptsize $\pm 5.22$} \\
        \nn{20} & 93.56 {\scriptsize $\pm 4.93$} \\
        \nn{1} & 92.81 {\scriptsize $\pm 5.69$} \\
        \aggmax & 92.81 {\scriptsize $\pm 5.69$} \\
        \nn{50} & 92.91 {\scriptsize $\pm 5.25$} \\
    \end{tabular}
    }
    \centering
    \raisebox{-2.5em}{ 
        \includegraphics[width=0.6\linewidth]{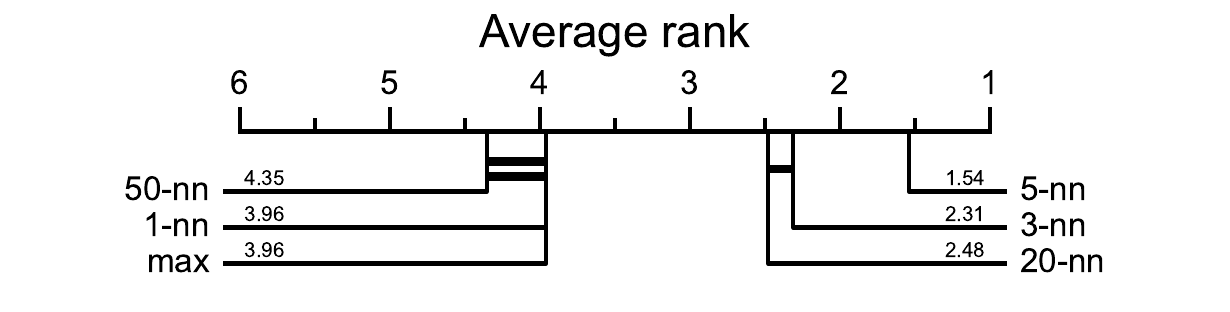}
    }
    \label{cd:agg-all}
\end{figure}

\begin{figure}[tbhp]
    \captionlistentry[table]{A table beside a figure}
    \captionsetup{labelformat=andfigure}
    \caption{Benchmarking \AGG mechanism across \model{s} and \task{s}. \GEN is fixed to \kmeans with $k=20$. \ll normalization is used for all \NORMs. The CD diagram shows that \aggmax is the best-performing aggregation method.}
    {
    \begin{tabular}{lr}
        \AGG & Avg. acc. \% {\scriptsize $\pm$ std} \\
        \hline
        \nn{1} & \textbf{91.06} {\scriptsize $\pm 6.21$} \\
        \aggmax & \textbf{91.06} {\scriptsize $\pm 6.21$} \\
        \nn{3} & 90.59 {\scriptsize $\pm 6.16$} \\
        \nn{5} & 90.12 {\scriptsize $\pm 6.21$} \\
        \nn{20} & 87.05 {\scriptsize $\pm 7.46$} \\
    \end{tabular}
    }
    \centering
    \raisebox{-3em}{ 
        \includegraphics[width=0.6\linewidth]{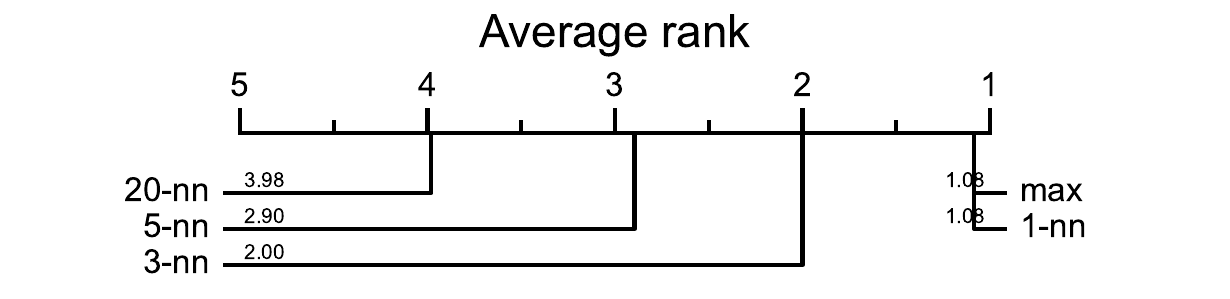}
    }
    \label{cd:agg-k20}
\end{figure}

\paragraph{Aggregation (\AGG).}
In addition to \aggmax, we study the effect of using $m$-nearest neighbor (\nn{m}) as an aggregation method.
Recall that \aggmax is a special case of \nn{m} when $m=1$ (as \NORMpost is set to \ll).
We investigate different values for $m \in \{1,3,5,20,50\}$.

When \all data is imprinted, \cref{cd:agg-all} shows that using \nn{m} aggregation for $m\in\{3,5,20\}$ is slightly better than \aggmax.
With $k=20$ and \kmeans as \GEN, \aggmax ($=$\nn{1}) aggregation becomes the top performing combination (see \cref{cd:agg-k20}).
Furthermore, the reduction of proxies (from \all ($\approx 6000$ per class) to $k=20$) leads to a decrease in accuracy of less than $3\%$.

\begin{figure}[htb]
    \centering
    \includegraphics[width=0.83\linewidth]{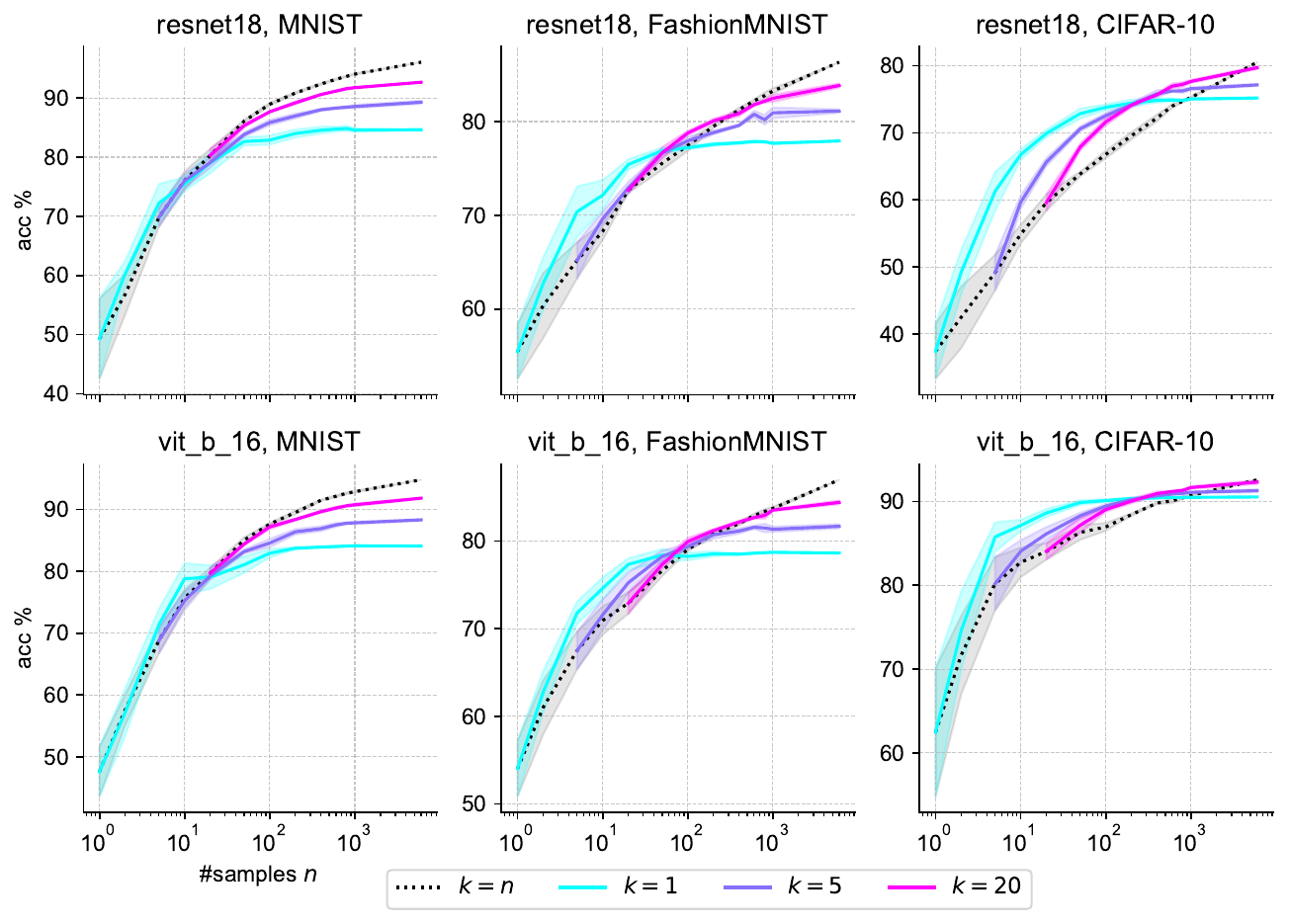}
    \caption{
        \kmeans with different values for $k$ in \fewshot{n} scenarios.
        Shaded areas indicate $95\%$ confidence intervals (CIs) over five different seeds.
        Other variables are fixed to our previously described best method of using \ll normalizations and \aggmax as \AGG.
        Note that only data for the meaningful case of $k \le n$ is shown.
        For \mnist and \fmnist, \mean ceases to be the best strategy already at only roughly 50 samples.
    }
    \label{fig:k-vs-n}
\end{figure}

\subsection{Low-Data Regime}\label{ss:lowdata}
We analyze the \fewshot{n} scenario with our method (\kmeans as \GEN, \ll as \NORMs, and \aggmax as \AGG).
Furthermore, we focus on the large tasks \task containing all ten classes of \mnist, \fmnist, resp. \cifar at once.
In this scenario, due to sampling only a few examples $n$, we average over five (instead of three) different seeds.
\par
From the results in \cref{fig:k-vs-n}, we find that as the number of samples $n$ increases, \kmeans starts to outperform \mean imprinting.
The use of more proxies $k$ further amplifies this performance gain.
The shift occurs at roughly 50 samples per class for \mnist and \fmnist, while for \cifar, $k>1$ becomes prominently better at around 200 samples per class (see \cref{fig:k-vs-small-n} for a display of all \model{s} focused on $10 \le n \le 400$).


\subsection{Neural Collapse and Number of Proxies}
\label{ss:nc-and-acc}

\begin{figure}[tbhp]
    \centering
    \begin{subfigure}[t]{0.42\linewidth}
        \vspace{0pt}
        \centering
        \includegraphics[width=\linewidth]{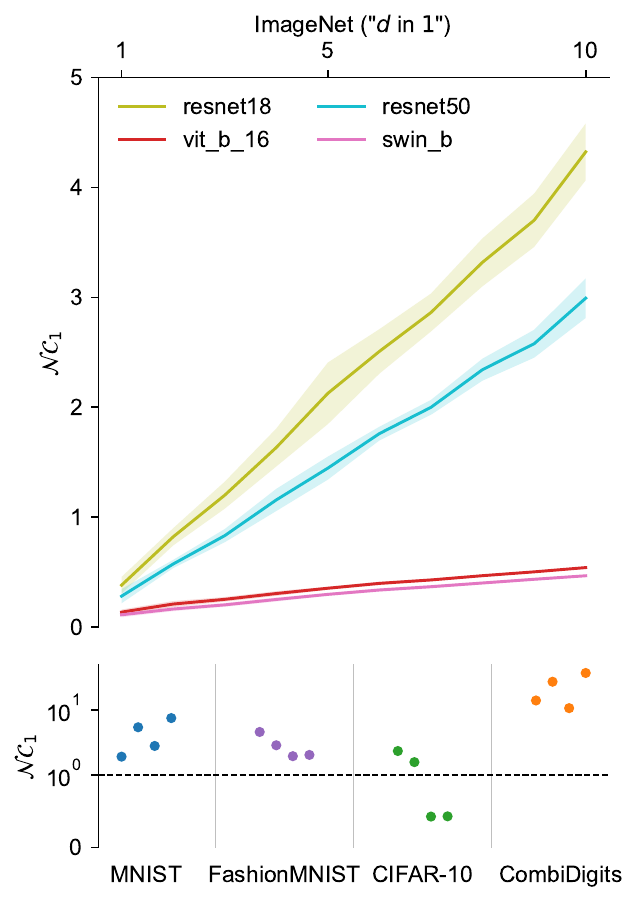}
        \caption{
        A clear linear relationship between the neural collapse measure \ncone and $d$ can be inferred for all \model{s}, i.e., increased multi-modality implies less collapse.
        The \ncone of other datasets (\combidigits in particular) is much higher across all \model{s}, while only \vitb and \swinb get an \ncone of less than one on \cifar.
        }
        \label{fig:nc1s}
    \end{subfigure}
    \hfill
    \begin{subfigure}[t]{0.54\linewidth}
        \vspace{0pt}
        \centering
        \includegraphics[width=0.85\linewidth]{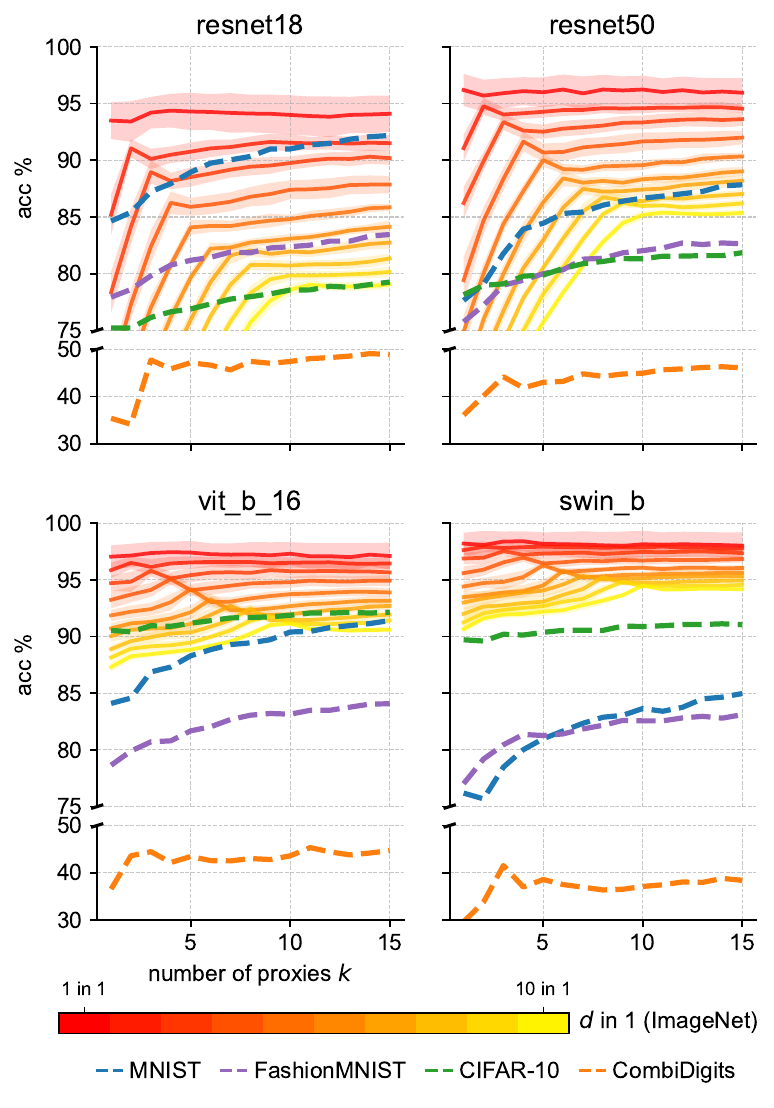}
        \caption{
        Accuracy over number of proxies $k$ used in \kmeans together with \ll for all \NORMs and \aggmax as aggregation.
        In all four subplots, peaks in accuracy at $k=d$ can be inferred.
        This confirms the connection between the effect of using multiple proxies and the collapse of the data.
        }
        \label{fig:label-remappings-task-acc-vs-num-proxies-qinorm}
    \end{subfigure}
    \caption{
        Averaged \ncone resp. accuracy of ten random \imagenet label remappings (\enquote{$d$ in $1$}) for every $d = 1,\dotsc,10$.
        Shaded areas indicate $95\%$ CIs over three different seeds.
        Values for the tasks containing all of \mnist, \fmnist, \cifar, resp. \combidigits at once are shown in dotted styles.
    }   
\end{figure}

\Cref{fig:nc1s} depicts the neural collapse measurement \ncone (see \cref{eq:ncone}) for the 100 random \imagenet tasks with remapped labels as explained in \cref{sec:experiments}, as well as the four tasks containing all of \mnist, \fmnist, \cifar, resp. \combidigits.
It can be inferred that \imagenet has a close-to-zero \ncone score, which increases linearly when adding more classes to each label (i.e., increasing multi-modality).
As for the other datasets, \cifar is generally more collapsed according to its low value of \ncone, which even falls below $1$ for the Transformer-based \model{s}.
We hypothesize that this is due to the similarity of its categories to those appearing in \imagenet.
By design, the synthetic \combidigits dataset, introduced in \cref{sec:experiments}, has a very high \ncone.
Apart from that, \ncone of the \imagenet data for the Transformer-based architectures are much lower and therefore they are more collapsed compared to the CNN-based \model{s}.
These architectural differences are further investigated in \cref{ss:archidifferences}.

For the same data, \cref{fig:label-remappings-task-acc-vs-num-proxies-qinorm} depicts accuracy over a varying number of proxies $k$ inferred from \kmeans.
A prominent peak at $k=d$ can be inferred for every \model and reflects that $d$ class proxies lead to the best result for $d$-modal class distributions.
Furthermore, increasing $k$ for the \imagenet sets has a much larger effect on the CNN-based \model{s}.
We argue that this is because of their higher values of \ncone, indicating less neural collapse.
In \cref{ss:archidifferences}, we analyze architectural differences and training setups to explain these variations. 
The fact that \cifar has the lowest \ncone scores (see \cref{fig:nc1s}) is reflected by flat green curves over $k$.

\begin{figure}[htbp]
    \centering
    \includegraphics[width=0.6\linewidth]{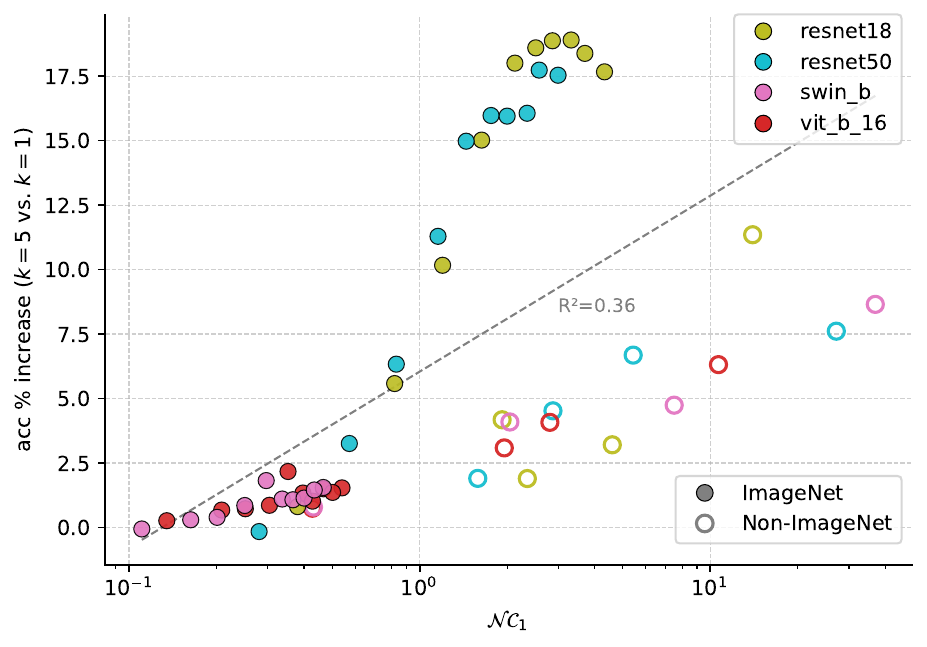}
    \caption{
        Accuracy (\%) gain when using $k=5$ instead of $k=1$ proxies per class with \kmeans, plotted against the neural collapse measure \ncone (log-scaled axis).
        Colors indicate different \model{s}.
        Outlined markers correspond to non-\imagenet datasets (\mnist, \fmnist, \cifar, and \combidigits), while filled markers denote averaged results from our random \imagenet label remappings.
        The dashed gray line shows a least-squares fit ($R^2 = 0.36$), summarizing an approximate log-linear dependence between \ncone\ and the gains from using $k>1$ proxies.
        Across all points, the Spearman rank correlation is $\rho = 0.82$ ($p<0.0001$), indicating a statistically significant positive association.
        Consistent with this trend, for regimes with $\mathcal{NC}_1 > 1$ single-proxy \mean imprinting ($k=1$) appears to become substantially suboptimal compared to using multiple proxies ($k>1$).
    }
    \label{fig:acc_inc_vs_nc1}
\end{figure}

\Cref{fig:acc_inc_vs_nc1} renders this connection more precise: For datasets with large \ncone, i.e., where intra-class variability surpasses inter-class variability, using more than a single proxy per class ($k>1$) yields a clear performance gain over mean imprinting ($k=1$).


\FloatBarrier

\section{Conclusion}\label{ss:conclusion}

We present a new framework, \framework, to analyze three main components relevant to weight imprinting, namely, weight generation, normalization, and aggregation.
Within \framework, state-of-the-art imprinting strategies become special cases.
This allows for a comprehensive analysis of different approaches through systematic experiments and leads us to generalize to a new, best-performing imprinting variant.
That is, using \kmeans weight generation with \ll normalizations and \aggmax aggregation, which outperforms all previously studied methods (see \cref{tab:special-cases}).

\paragraph{\kmeans generates better weights than \mean.}
In particular, we find that the \mean weight generation (\GEN) method, despite its prominence in previous work, falls short compared to \kmeans{} -- even when the number of proxies $k$ is very small.
Remarkably, with as little as 50 samples per class, \kmeans can already outperform the original imprinting method proposed by \citet{qi-imprinting}, highlighting its advantage in low-data regimes.

\paragraph{\ll weight normalization is essential for strong performance.}
The \aggmax aggregation directly scales with the magnitude of the weights.
Normalization (\NORMpost) ensures that all proxies have the same magnitude, preventing differences in vector norms from disproportionately affecting classifier predictions.
Nearest neighbor (\nn{l}) aggregation is not as affected by the lack of normalization, since it uses Euclidean distance.
Although still part of common procedure, normalizations for embeddings (\NORMpre and \NORMinf) appear to have minimal impact on performance.

\paragraph{With \aggmax aggregation, there is no need to store \all data.}
While nearest neighbor (\nn{m}) aggregation (\AGG) performs well when all data is saved (e.g., when there are no storage constraints), \aggmax aggregation with limited number of representative proxies (e.g., through $k$-means) is an efficient alternative without a substantial loss in performance.

\paragraph{Neural collapse proves the efficacy of imprinting.}
During training, the last-layer weights of a \model tend to collapse to their respective class means.
This proves the success of \mean imprinting on known classes.
New, out-of-distribution data, however, often shows less collapse, making it beneficial to imprint more than one proxy.
Our results show that the advantage of using multiple proxies is strongly coupled to the degree of neural collapse: the accuracy gain from using $k>1$ proxies increases approximately log-linearly with \ncone.
In particular, multi-proxy imprinting yields substantial and predictable improvements on datasets with \ncone exceeding 1.
While from a practical perspective, choosing the exact number of proxies $k$ based on pure greedy search with validation data (or as part of any AutoML pipeline) is still a valid option, our analysis provides insights into the underlying mechanism, turning \ncone into an indicator for when multi-proxy imprinting is highly benefitial.

\paragraph{Limitations.}
Our experiments are limited to foundation models for image classification and do not cover other data modalities such as audio, text, or sensor inputs.
While \framework is agnostic to modality, empirical validation of our results outside of vision is needed.
Although imprinting alone provides an efficient solution to transfer learning, when compared to purely gradient-based learning, a gap still remains.
We do not investigate the benefit of combining it with optimization methods such as gradient-based learning, and the choice of $k$ still requires heuristic or empirical selection rather than direct prediction.

\paragraph{Future Work.}
The usage of both weight and activation sparsity as \citet{shen2023reducing} could change the within- and between-class variability in favor of using a higher number of proxies.
Besides only using the penultimate layer embeddings for generating the classifier weights, an interesting area of study could be extracting embeddings from previous layers of the \model for this purpose.
Recently, the study by \citet{marczak2025revisiting} showed that adding a multi-layer perceptron projector between the penultimate and classification layers results in representations that are more transferable.
Another avenue of research is to thoroughly analyze imprinting the weights of other layers as well \citep{masked-imprinting}.


{\small
\subsubsection*{Acknowledgments}
Our work is funded by the Deutsche Forschungsgemeinschaft (DFG, German Research Foundation) -- FIP-12 -- Project-ID 528483508, as well as the European Union under the grant project 101079894 (COMFORT - Improving Urologic Cancer Care with Artificial Intelligence Solutions). Views and opinions expressed are however those of the author(s) only and do not necessarily reflect those of the European Union or European Health and Digital Executive Agency (HADEA). Neither the European Union nor the granting authority can be held responsible for them.
We thank Viet Anh Khoa Tran for initial discussions about the neural collapse phenomenon.
We further thank the reviewers and the Action Editor for their detailed and constructive feedback, which substantially improved this work.

\subsubsection*{Author Contributions}
JW contributed to the development of the framework, conducting experiments and evaluated the findings.
GA was responsible for investigating NC measures and overall contribution to the project.
MK contributed to extending the framework and handling data preparation.
AF provided critical feedback on the presentation of the results and contributed to refining the manuscript.
AL, ER, and FG provided supervision, contributed to the overall concepts presented, and to refining the manuscript.

}

\clearpage
{
\small
\bibliographystyle{tmlr}
\bibliography{main}

@inproceedings{qi-imprinting,
  title={Low-shot learning with imprinted weights},
  author={Qi, Hang and Brown, Matthew and Lowe, David G},
  booktitle={Proceedings of the IEEE conference on computer vision and pattern recognition},
  pages={5822--5830},
  year={2018}
}

@article{usps,
  title={A database for handwritten text recognition research},
  author={Hull, Jonathan J.},
  journal={IEEE Transactions on pattern analysis and machine intelligence},
  volume={16},
  number={5},
  pages={550--554},
  year={1994},
  publisher={IEEE}
}

@inproceedings{svhn,
  title={Reading digits in natural images with unsupervised feature learning},
  author={Netzer, Yuval and Wang, Tao and Coates, Adam and Bissacco, Alessandro and Wu, Baolin and Ng, Andrew Y and others},
  booktitle={NIPS workshop on deep learning and unsupervised feature learning},
  volume={2011},
  pages={4},
  year={2011},
  organization={Granada}
}

@article{mnistm,
  title={Domain-adversarial training of neural networks},
  author={Ganin, Yaroslav and Ustinova, Evgeniya and Ajakan, Hana and Germain, Pascal and Larochelle, Hugo and Laviolette, Fran{\c{c}}ois and March, Mario and Lempitsky, Victor},
  journal={Journal of machine learning research},
  volume={17},
  number={59},
  pages={1--35},
  year={2016}
}

@article{harun2025good,
  title={A Good Start Matters: Enhancing Continual Learning with Data-Driven Weight Initialization},
  author={Harun, Md Yousuf and Kanan, Christopher},
  journal={arXiv preprint arXiv:2503.06385},
  year={2025}
}

@misc{cifar10,
  title        = {{CIFAR-10}},
  year = {2009},
  author = {Krizhevsky, Alex and Hinton, Geoffrey and others},
 publisher={Toronto, ON, Canada},
  howpublished = {\url{https://www.cs.toronto.edu/~kriz/cifar.html}},
}

@misc{coralai,
    title = {Retrain a classification model on-device with weight imprinting},
    author = {Google Coral},
    howpublished = {\url{https://coral.ai/docs/edgetpu/retrain-classification-ondevice}}
}

@article{transferlearning2,
  title={How transferable are features in deep neural networks?},
  author={Yosinski, Jason and Clune, Jeff and Bengio, Yoshua and Lipson, Hod},
  journal={Advances in neural information processing systems},
  volume={27},
  year={2014}
}

@article{scikit-learn,
  title={Scikit-learn: Machine Learning in {P}ython},
  author={Pedregosa, F. and Varoquaux, G. and Gramfort, A. and Michel, V.
          and Thirion, B. and Grisel, O. and Blondel, M. and Prettenhofer, P.
          and Weiss, R. and Dubourg, V. and Vanderplas, J. and Passos, A. and
          Cournapeau, D. and Brucher, M. and Perrot, M. and Duchesnay, E.},
  journal={Journal of Machine Learning Research},
  volume={12},
  pages={2825--2830},
  year={2011}
}

@article{holm1979simple,
  title={A simple sequentially rejective multiple test procedure},
  author={Holm, Sture},
  journal={Scandinavian journal of statistics},
  pages={65--70},
  year={1979},
  publisher={JSTOR}
}

@inproceedings{xlstm,
  author       = {Maximilian Beck and
                  Korbinian P{\"{o}}ppel and
                  Markus Spanring and
                  Andreas Auer and
                  Oleksandra Prudnikova and
                  Michael Kopp and
                  G{\"{u}}nter Klambauer and
                  Johannes Brandstetter and
                  Sepp Hochreiter},
  editor       = {Amir Globersons and
                  Lester Mackey and
                  Danielle Belgrave and
                  Angela Fan and
                  Ulrich Paquet and
                  Jakub M. Tomczak and
                  Cheng Zhang},
  title        = {xLSTM: Extended Long Short-Term Memory},
  booktitle    = {Advances in Neural Information Processing Systems 38: Annual Conference
                  on Neural Information Processing Systems 2024, NeurIPS 2024, Vancouver,
                  BC, Canada, December 10 - 15, 2024},
  year         = {2024},
  url          = {http://papers.nips.cc/paper\_files/paper/2024/hash/c2ce2f2701c10a2b2f2ea0bfa43cfaa3-Abstract-Conference.html},
  bibsource    = {dblp computer science bibliography, https://dblp.org}
}

@article{anderson77,
  title={Distinctive features, categorical perception, and probability learning: Some applications of a neural model.},
  author={Anderson, James A and Silverstein, Jack W and Ritz, Stephen A and Jones, Randall S},
  journal={Psychological review},
  volume={84},
  number={5},
  pages={413},
  year={1977},
  publisher={American Psychological Association}
}

@article{kohonen,
  title={Correlation matrix memories},
  author={Kohonen, Teuvo},
  journal={IEEE transactions on computers},
  volume={100},
  number={4},
  pages={353--359},
  year={2009},
  publisher={IEEE}
}

@article{anderson72,
  title={A simple neural network generating an interactive memory},
  author={Anderson, James A},
  journal={Mathematical biosciences},
  volume={14},
  number={3-4},
  pages={197--220},
  year={1972},
  publisher={Elsevier}
}

@article{imprintingrobot,
  title={Weight imprinting classification-based force grasping with a variable-stiffness robotic gripper},
  author={Zhu, Haiyue and Li, Xiong and Chen, Wenjie and Li, Xiaocong and Ma, Jun and Teo, Chek Sing and Teo, Tat Joo and Lin, Wei},
  journal={IEEE Transactions on Automation Science and Engineering},
  volume={19},
  number={2},
  pages={969--981},
  year={2022},
  publisher={IEEE}
}

@article{nakano,
  title={Associatron-a model of associative memory},
  author={Nakano, Kaoru},
  journal={IEEE Transactions on systems, man, and cybernetics},
  pages={380--388},
  year={2007},
  publisher={IEEE}
}

@article{dayan,
  title={Optimising synaptic learning rules in linear associative memories},
  author={Dayan, Peter and Willshaw, David J},
  journal={Biological cybernetics},
  volume={65},
  number={4},
  pages={253--265},
  year={1991},
  publisher={Springer}
}

@article{sejnowski,
  title={Storing covariance with nonlinearly interacting neurons},
  author={Sejnowski, Terrence J},
  journal={Journal of mathematical biology},
  volume={4},
  number={4},
  pages={303--321},
  year={1977},
  publisher={Springer}
}

@inproceedings{metric-proxies,
  title={No fuss distance metric learning using proxies},
  author={Movshovitz-Attias, Yair and Toshev, Alexander and Leung, Thomas K and Ioffe, Sergey and Singh, Saurabh},
  booktitle={Proceedings of the IEEE international conference on computer vision},
  pages={360--368},
  year={2017}
}

@article{first-nearestclassmean,
  title={Distance-based image classification: Generalizing to new classes at near-zero cost},
  author={Mensink, Thomas and Verbeek, Jakob and Perronnin, Florent and Csurka, Gabriela},
  journal={IEEE transactions on pattern analysis and machine intelligence},
  volume={35},
  number={11},
  pages={2624--2637},
  year={2013},
  publisher={IEEE}
}

@article{prototypical,
  title={Prototypical networks for few-shot learning},
  author={Snell, Jake and Swersky, Kevin and Zemel, Richard},
  journal={Advances in neural information processing systems},
  volume={30},
  year={2017}
}

@article{hypersphere-imprinting,
  title={Hypersphere-based weight imprinting for few-shot learning on embedded devices},
  author={Passalis, Nikolaos and Iosifidis, Alexandros and Gabbouj, Moncef and Tefas, Anastasios},
  journal={IEEE Transactions on Neural Networks and Learning Systems},
  volume={32},
  number={2},
  pages={925--930},
  year={2020},
  publisher={IEEE}
}

@inproceedings{multiple-projection-head-imprinting,
  title={Few shot model based on weight imprinting with multiple projection head},
  author={Cristovao, Paulino and Nakada, Hidemoto and Tanimura, Yusuke and Asoh, Hideki},
  booktitle={2022 16th International Conference on Ubiquitous Information Management and Communication (IMCOM)},
  pages={1--7},
  year={2022},
  organization={IEEE}
}

@article{zhu2021geometric,
  title={A geometric analysis of neural collapse with unconstrained features},
  author={Zhu, Zhihui and Ding, Tianyu and Zhou, Jinxin and Li, Xiao and You, Chong and Sulam, Jeremias and Qu, Qing},
  journal={Advances in Neural Information Processing Systems},
  volume={34},
  pages={29820--29834},
  year={2021}
}

@inproceedings{imprint-objdetec,
  title={Classification weight imprinting for data efficient object detection},
  author={Li, Yiting and Zhu, Haiyue and Ma, Jun and Tian, Sichao and Teo, Chek Sing and Xiang, Cheng and Vadakkepa, Prahlad and Lee, Tong Heng},
  booktitle={2021 IEEE 30th International Symposium on Industrial Electronics (ISIE)},
  pages={1--5},
  year={2021},
  organization={IEEE}
}

@inproceedings{transferlearning3,
  title={Deep learning of representations for unsupervised and transfer learning},
  author={Bengio, Yoshua},
  booktitle={Proceedings of ICML workshop on unsupervised and transfer learning},
  pages={17--36},
  year={2012},
  organization={JMLR Workshop and Conference Proceedings}
}

@article{cddiag-code,
  title={Deep learning for time series classification: a review},
  author={Ismail Fawaz, Hassan and Forestier, Germain and Weber, Jonathan and Idoumghar, Lhassane and Muller, Pierre-Alain},
  journal={Data mining and knowledge discovery},
  volume={33},
  number={4},
  pages={917--963},
  year={2019},
  publisher={Springer}
}

@inproceedings{DBLP:conf/icml/TirerHN23,
  title={Perturbation analysis of neural collapse},
  author={Tirer, Tom and Huang, Haoxiang and Niles-Weed, Jonathan},
  booktitle={International Conference on Machine Learning},
  pages={34301--34329},
  year={2023},
  organization={PMLR}
}

@article{belal2025fsid,
  title={FSID: a novel approach to human activity recognition using few-shot weight imprinting},
  author={Belal, Mohammad and Hassan, Taimur and Hassan, Abdelfatah and Velayudhan, Divya and Elhendawi, Noureldin and Aljarah, Ahmad and Hussain, Irfan},
  journal={Scientific Reports},
  volume={15},
  number={1},
  pages={20865},
  year={2025},
  publisher={Nature Publishing Group UK London}
}

@inproceedings{icarl,
  title={icarl: Incremental classifier and representation learning},
  author={Rebuffi, Sylvestre-Alvise and Kolesnikov, Alexander and Sperl, Georg and Lampert, Christoph H},
  booktitle={Proceedings of the IEEE conference on Computer Vision and Pattern Recognition},
  pages={2001--2010},
  year={2017}
}

@inproceedings{
    DBLP:conf/iclr/Galanti0H22,
    title={On the Role of Neural Collapse in Transfer Learning},
    author={Tomer Galanti and Andr{\'a}s Gy{\"o}rgy and Marcus Hutter},
    booktitle={International Conference on Learning Representations},
    year={2022}
}

@inproceedings{neuralcollapse2,
  author       = {X. Y. Han and
                  Vardan Papyan and
                  David L. Donoho},
  title        = {Neural Collapse Under {MSE} Loss: Proximity to and Dynamics on the
                  Central Path},
  booktitle    = {The Tenth International Conference on Learning Representations, {ICLR}
                  },
  year         = {2022},
}

@article{
    neuralcollapse3,
    title={Neural Collapse: A Review on Modelling Principles and Generalization},
    author={Vignesh Kothapalli},
    journal={Transactions on Machine Learning Research},
    issn={2835-8856},
    year={2023}
}

@article{covidimprint,
  title={COVID-19 detection from chest x-ray images using imprinted weights approach},
  author={Zhang, Jianxing and Xi, Pengcheng and Ebadi, Ashkan and Azimi, Hilda and Tremblay, St{\'e}phane and Wong, Alexander},
  journal={arXiv preprint arXiv:2105.01710},
  year={2021}
}

@article{imprintallknn,
  title={Online continual learning without the storage constraint},
  author={Prabhu, Ameya and Cai, Zhipeng and Dokania, Puneet and Torr, Philip and Koltun, Vladlen and Sener, Ozan},
  journal={arXiv preprint arXiv:2305.09253},
  year={2023}
}

@article{shen2023reducing,
  title={Reducing catastrophic forgetting with associative learning: a lesson from fruit flies},
  author={Shen, Yang and Dasgupta, Sanjoy and Navlakha, Saket},
  journal={Neural Computation},
  volume={35},
  number={11},
  pages={1797--1819},
  year={2023},
  publisher={MIT Press One Rogers Street, Cambridge, MA 02142-1209, USA journals-info~…}
}

@article{yan2023few,
  title={Few-shot object detection with weight imprinting},
  author={Yan, Dingtian and Huang, Jitao and Sun, Hai and Ding, Fuqiang},
  journal={Cognitive Computation},
  volume={15},
  number={5},
  pages={1725--1735},
  year={2023},
  publisher={Springer}
}

@inproceedings{masked-imprinting,
  title={Amp: Adaptive masked proxies for few-shot segmentation},
  author={Siam, Mennatullah and Oreshkin, Boris N and Jagersand, Martin},
  booktitle={Proceedings of the IEEE/CVF International Conference on Computer Vision},
  pages={5249--5258},
  year={2019}
}

@article{foundationmodels,
  title={On the opportunities and risks of foundation models},
  author={Bommasani, Rishi and Hudson, Drew A and Adeli, Ehsan and Altman, Russ and Arora, Simran and von Arx, Sydney and Bernstein, Michael S and Bohg, Jeannette and Bosselut, Antoine and Brunskill, Emma and others},
  journal={arXiv preprint arXiv:2108.07258},
  year={2021}
}

@article{papyan2020prevalence,
  title={Prevalence of neural collapse during the terminal phase of deep learning training},
  author={Papyan, Vardan and Han, XY and Donoho, David L},
  journal={Proceedings of the National Academy of Sciences},
  volume={117},
  number={40},
  pages={24652--24663},
  year={2020},
  publisher={National Acad Sciences}
}

@article{fashionmnist,
  title={Fashion-mnist: a novel image dataset for benchmarking machine learning algorithms},
  author={Xiao, Han and Rasul, Kashif and Vollgraf, Roland},
  journal={arXiv preprint arXiv:1708.07747},
  year={2017}
}

@article{mnist,
  title={The mnist database of handwritten digit images for machine learning research [best of the web]},
  author={Deng, Li},
  journal={IEEE signal processing magazine},
  volume={29},
  number={6},
  pages={141--142},
  year={2012},
  publisher={IEEE}
}

@inproceedings{swin,
  title={Swin transformer: Hierarchical vision transformer using shifted windows},
  author={Liu, Ze and Lin, Yutong and Cao, Yue and Hu, Han and Wei, Yixuan and Zhang, Zheng and Lin, Stephen and Guo, Baining},
  booktitle={Proceedings of the IEEE/CVF international conference on computer vision},
  pages={10012--10022},
  year={2021}
}

@inproceedings{resnet,
  title={Deep residual learning for image recognition},
  author={He, Kaiming and Zhang, Xiangyu and Ren, Shaoqing and Sun, Jian},
  booktitle={Proceedings of the IEEE conference on computer vision and pattern recognition},
  pages={770--778},
  year={2016}
}

@article{cddiag,
  title={Statistical comparisons of classifiers over multiple data sets},
  author={Dem{\v{s}}ar, Janez},
  journal={The Journal of Machine learning research},
  volume={7},
  pages={1--30},
  year={2006},
  publisher={JMLR. org}
}

@inproceedings{
vit,
title={An Image is Worth 16x16 Words: Transformers for Image Recognition at Scale},
author={Alexey Dosovitskiy and Lucas Beyer and Alexander Kolesnikov and Dirk Weissenborn and Xiaohua Zhai and Thomas Unterthiner and Mostafa Dehghani and Matthias Minderer and Georg Heigold and Sylvain Gelly and others},
booktitle={International Conference on Learning Representations},
year={2021},
}

@inproceedings{imagenet,
  title={Imagenet: A large-scale hierarchical image database},
  author={Deng, Jia and Dong, Wei and Socher, Richard and Li, Li-Jia and Li, Kai and Fei-Fei, Li},
  booktitle={2009 IEEE conference on computer vision and pattern recognition},
  pages={248--255},
  year={2009},
}

@article{quantnorm1,
  title={Analysis of data from viral DNA microchips},
  author={Amaratunga, Dhammika and Cabrera, Javier},
  journal={Journal of the American Statistical Association},
  volume={96},
  number={456},
  pages={1161--1170},
  year={2001},
  publisher={Taylor \& Francis}
}

@article{quantnorm2,
  title={A comparison of normalization methods for high density oligonucleotide array data based on variance and bias},
  author={Bolstad, Benjamin M and Irizarry, Rafael A and {\AA}strand, Magnus and Speed, Terence P.},
  journal={Bioinformatics},
  volume={19},
  number={2},
  pages={185--193},
  year={2003},
  publisher={Oxford University Press}
}

@inproceedings{convolutional-prototypes,
  title={Robust classification with convolutional prototype learning},
  author={Yang, Hong-Ming and Zhang, Xu-Yao and Yin, Fei and Liu, Cheng-Lin},
  booktitle={Proceedings of the IEEE conference on computer vision and pattern recognition},
  pages={3474--3482},
  year={2018}
}

@inproceedings{semanticdriftCL,
  title={Semantic drift compensation for class-incremental learning},
  author={Yu, Lu and Twardowski, Bartlomiej and Liu, Xialei and Herranz, Luis and Wang, Kai and Cheng, Yongmei and Jui, Shangling and Weijer, Joost van de},
  booktitle={Proceedings of the IEEE/CVF conference on computer vision and pattern recognition},
  pages={6982--6991},
  year={2020}
}

@article{imprint-multilabel,
  title={Personalizing pre-trained models},
  author={Khan, Mina and Srivatsa, P and Rane, Advait and Chenniappa, Shriram and Hazariwala, Asadali and Maes, Pattie},
  journal={arXiv preprint arXiv:2106.01499},
  year={2021}
}

@inproceedings{fewshot-metagenweights,
  title={Dynamic few-shot visual learning without forgetting},
  author={Gidaris, Spyros and Komodakis, Nikos},
  booktitle={Proceedings of the IEEE conference on computer vision and pattern recognition},
  pages={4367--4375},
  year={2018}
}

@inproceedings{
    ncmbaseline,
    title={A Simple Baseline that Questions the Use of Pretrained-Models in Continual Learning},
    author={Paul Janson and Wenxuan Zhang and Rahaf Aljundi and Mohamed Elhoseiny},
    booktitle={NeurIPS 2022 Workshop on Distribution Shifts: Connecting Methods and Applications},
    year={2022}
}

@article{itsdone,
  title={A single fast Hebbian-like process enabling one-shot class addition in deep neural networks without backbone modification},
  author={Hosoda, Kazufumi and Nishida, Keigo and Seno, Shigeto and Mashita, Tomohiro and Kashioka, Hideki and Ohzawa, Izumi},
  journal={Frontiers in Neuroscience},
  volume={18},
  pages={1344114},
  year={2024},
  publisher={Frontiers Media SA}
}

@inproceedings{marczak2025revisiting,
  title={Revisiting Supervision for Continual Representation Learning},
  author={Marczak, Daniel and Cygert, Sebastian and Trzci{\'n}ski, Tomasz and Twardowski, Bart{\l}omiej},
  booktitle={European Conference on Computer Vision},
  pages={181--197},
  year={2025},
  organization={Springer}
}

@inproceedings{devlin_bert_2019,
	title = {{BERT}: {Pre}-training of {Deep} {Bidirectional} {Transformers} for {Language} {Understanding}},
	booktitle = {{NAACL}-{HLT}},
	author = {Devlin, Jacob and Chang, Ming-Wei and Lee, Kenton and Toutanova, Kristina},
	year = {2019},
	pages = {4171--4186},
}

@inproceedings{donahue_decaf_2014,
	title = {{DeCAF}: {A} {Deep} {Convolutional} {Activation} {Feature}  for {Generic} {Visual} {Recognition}},
	author = {Donahue, Jeff and Jia, Yangqing and Vinyals, Oriol and Hoffman, Judy and Zhang, Ning and Tzeng, Eric and Darrell, Trevor},
	year = {2014},
	pages = {647--655},
	booktitle = {ICML}
}

@inproceedings{kornblith_better_2019,
	address = {Long Beach, CA, USA},
	title = {Do {Better} {ImageNet} {Models} {Transfer} {Better}?},
	booktitle = {CVPR},
	publisher = {IEEE},
	author = {Kornblith, Simon and Shlens, Jonathon and Le, Quoc V.},
	year = {2019},
	pages = {2656--2666}
}

@misc{huh_what_2016,
	title = {What makes {ImageNet} good for transfer learning?},
	author = {Huh, Minyoung and Agrawal, Pulkit and Efros, Alexei A.},
	year = {2016},
	note = {arXiv:1608.08614}
}
}

\clearpage
\appendix

\renewcommand{\theequation}{A.\arabic{equation}}
\renewcommand{\thefigure}{A.\arabic{figure}}
\renewcommand{\thetable}{A.\arabic{table}}

\setcounter{equation}{0}
\setcounter{figure}{0}
\setcounter{table}{0}

\section{Appendix}\label{appendix}

\subsection{Additional Results}
We provide additional tables and critical difference (CD) diagrams that expand on the results of \cref{sec:results}.

\begin{figure}[tbhp]
    \captionlistentry[table]{A table beside a figure}
    \captionsetup{labelformat=andfigure}
    \caption{Benchmarking \GEN mechanism for $k \le 5$ across \model{s} and \task{s}. Best \NORMs combination for each row is used implicitly. \AGG is fixed to \aggmax. CD diagram depicts statistical significance of \kmeans as \GEN. See \cref{cd:proxyanalysis} for $k \le 20$.}
    {\small
    \begin{tabular}{llr}
        \GEN & $k$ & Avg. acc. \% {\scriptsize $\pm$ std} \\
        \midrule
        \kmeans & 5 & \textbf{89.15} {\scriptsize $\pm 6.96$} \\
        \mean & 1 & 86.84 {\scriptsize $\pm 7.80$} \\
        \kmedoids & 5 & 84.87 {\scriptsize $\pm 8.76$} \\
        \kcovmax & 5 & 82.11 {\scriptsize $\pm 10.39$} \\
        \krandom & 5 & 75.93 {\scriptsize $\pm 11.92$} \\
        \kfps & 5 & 63.64 {\scriptsize $\pm 12.29$} \\ 
    \end{tabular}
    }
    \centering
    \raisebox{-2em}{ 
        \includegraphics[width=0.5\linewidth]{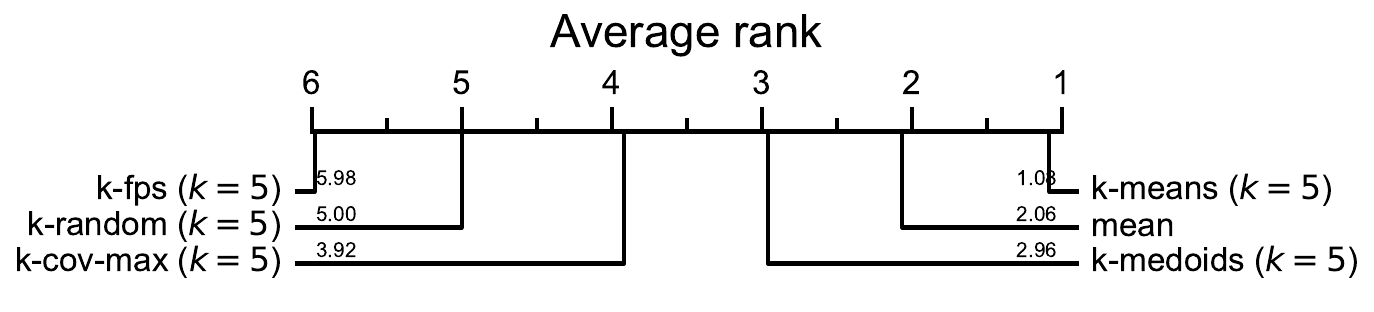}
    }
    \label{cd:proxyanalysis-less}
\end{figure}

\begin{figure}[tbhp]
    \captionlistentry[table]{A table beside a figure}
    \captionsetup{labelformat=andfigure}
    \caption{Benchmarking \NORMs across \model{s} and \task{s} shows crucial effect of \ll normalization. \GEN is fixed to \mean and \AGG to \aggmax. CD diagram depicting statistical significance of \ll for \NORMpost. Combinations are listed as \enquote{\NORMinf \& \NORMpre \& \NORMpost{}}.}
    {\small
    \begin{tabular}{lllr}
        \NORMinf & \NORMpre & \NORMpost & Avg. acc. \% {\scriptsize $\pm$ std} \\
        \midrule
        \ll & \none & \ll & \textbf{86.84} {\scriptsize $\pm 7.80$} \\
        \none & \none & \ll & \textbf{86.84} {\scriptsize $\pm 7.80$} \\
        \ll & \ll & \ll & \textbf{86.79} {\scriptsize $\pm 7.83$} \\
        \ll & \none & \quant & 82.90 {\scriptsize $\pm 12.87$} \\
        \none & \none & \quant & 82.90 {\scriptsize $\pm 12.87$} \\
        \ll & \ll & \quant & 82.83 {\scriptsize $\pm 12.86$} \\
        \ll & \ll & \none & 83.26 {\scriptsize $\pm 9.18$} \\
        \ll & \none & \none & 67.66 {\scriptsize $\pm 22.12$} \\
        \none & \none & \none & 67.66 {\scriptsize $\pm 22.12$} \\
    \end{tabular}
    }
    \centering
    \raisebox{-2em}{ 
        \includegraphics[width=0.5\linewidth]{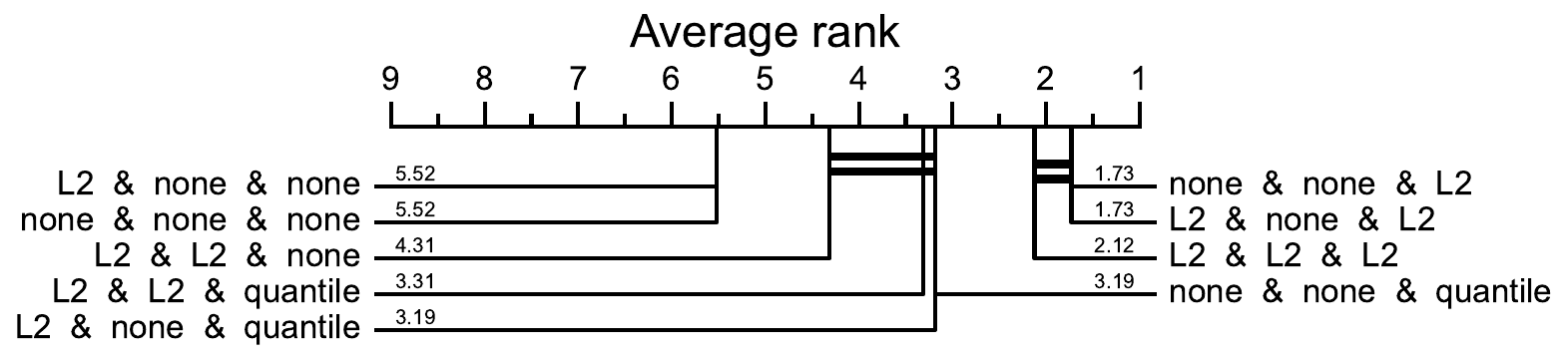}
    }
    \label{cd:normk1-full}
\end{figure}



\begin{figure}[tbhp]
    \captionlistentry[table]{A table beside a figure}
    \captionsetup{labelformat=andfigure}
    \caption{Benchmarking \NORMs across \model{s} and \task{s}. \GEN is fixed to \kmeans with $k=20$ and \AGG to \aggmax. CD diagram depicting statistical significance of \ll for \NORMpost. Combinations are listed as \enquote{\NORMinf \& \NORMpre \& \NORMpost.}}
    {\small
    \begin{tabular}{lllr}
        \NORMinf & \NORMpre & \NORMpost & Avg. acc. \% {\scriptsize $\pm$ std} \\
        \midrule
        \ll & \none & \ll & \textbf{91.04} {\scriptsize $\pm 6.26$} \\
        \none & \none & \ll & \textbf{91.04} {\scriptsize $\pm 6.26$} \\
        \ll & \ll & \ll & \textbf{91.06} {\scriptsize $\pm 6.21$} \\
        \ll & \ll & \quant & 90.51 {\scriptsize $\pm 6.40$} \\
        \ll & \ll & \none & 89.55 {\scriptsize $\pm 6.70$} \\
        \ll & \none & \quant & 79.21 {\scriptsize $\pm 15.20$} \\
        \none & \none & \quant & 79.21 {\scriptsize $\pm 15.20$} \\
        \ll & \none & \none & 73.53 {\scriptsize $\pm 21.05$} \\
        \none & \none & \none & 73.53 {\scriptsize $\pm 21.05$} \\
    \end{tabular}
    }
    \centering
    \raisebox{-2em}{ 
        \includegraphics[width=0.5\linewidth]{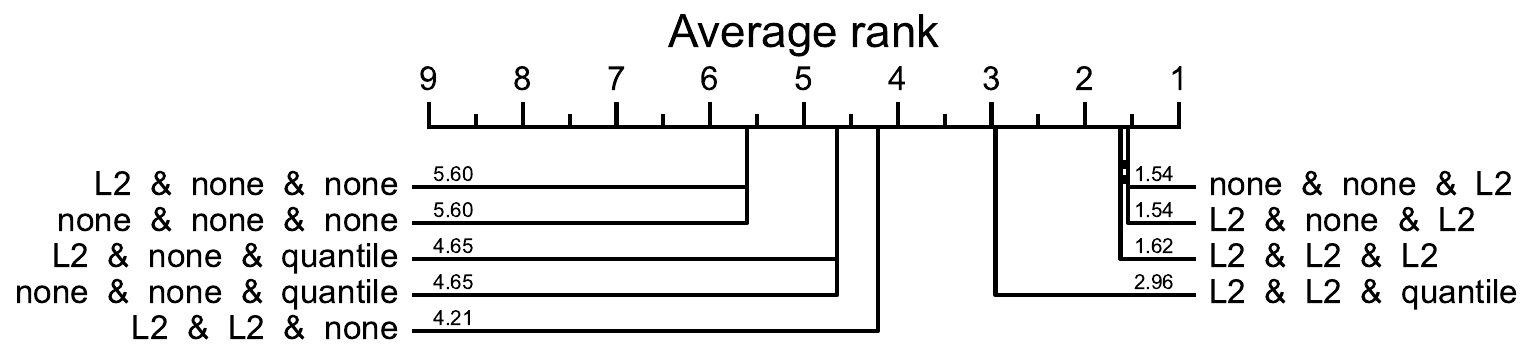}
    }
    \label{cd:normk20}
\end{figure}



\begin{figure}[tbhp]
    \centering
    \includegraphics[width=0.8\linewidth]{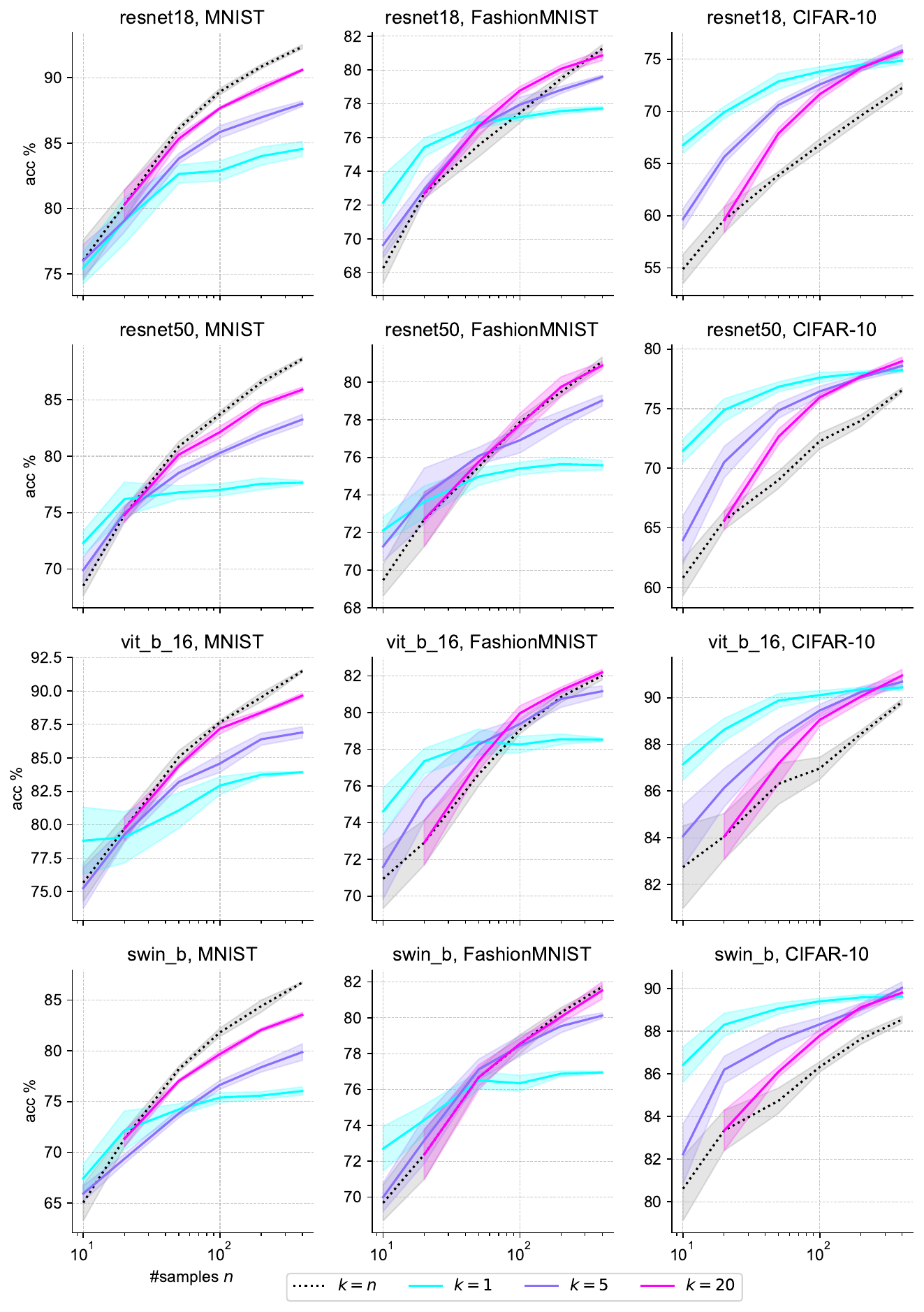}
    \caption{
        \kmeans with different values for $k$ in \fewshot{n} scenarios with focus on $10 \le n \le 400$.
        Shaded areas indicate $95\%$ confidence intervals (CIs) over five different seeds.
        Other variables are fixed to our previously described best method of using \ll for \NORMs and \aggmax as \AGG. Note that only data for the meaningful case of $k \le n$ is shown.
        For \mnist and \fmnist, \mean ceases to be the best strategy already at only roughly 50 samples.
        See \cref{fig:k-vs-n} for more values of $n$.}
    \label{fig:k-vs-small-n}
\end{figure}

\subsection{Datasets}\label{ss:datasets}
We briefly describe the datasets used in our experiments.

\paragraph{\imagenet~\citep{imagenet}.}
We use the ILSVRC 2012 version (commonly called \imagenet{-1K}) containing \numprint{1000} classes and 1.2M training images. Since the test set is not publicly available, we use the validation set (\numprint{50000} images) as a stand-in.
For neural collapse investigations, we construct tasks by randomly grouping $d$ classes into one label, producing “$d$ in 1” tasks as explained in \cref{sec:experiments}.

\paragraph{\mnist~\citep{mnist}.}
A benchmark dataset of handwritten digits (0–9), consisting of \numprint{60000} training and \numprint{10000} test grayscale images of size $28 \times 28$.

\paragraph{\fmnist~\citep{fashionmnist}.}
A drop-in replacement for \mnist with the same format and number of samples, containing grayscale images of fashion items across 10 classes.

\paragraph{\cifar~\citep{cifar10}.}
A dataset of $32 \times 32$ RGB images covering 10 object classes, with \numprint{50000} training and \numprint{10000} test samples.

\paragraph{\mnistm~\citep{mnistm}, \svhn~\citep{svhn}, \usps~\citep{usps}.}
Digit classification datasets each containing digits 0-9 with domain-specific visual characteristics.
\mnistm applies color and texture transformations to MNIST digits, yielding \numprint{60000} training and \numprint{10000} RGB images of size $28 \times 28$.
\svhn consists of digit crops from house numbers in Google Street View, totaling \numprint{73257} training and \numprint{26032} test images of $32 \times 32$ in RGB.
\usps contains scanned and normalized handwritten digits in $16 \times 16$ grayscale, split into \numprint{7219} training and \numprint{2007} test images.

\begin{figure}[tbhp]
    \centering
    \includegraphics[width=0.4\linewidth]{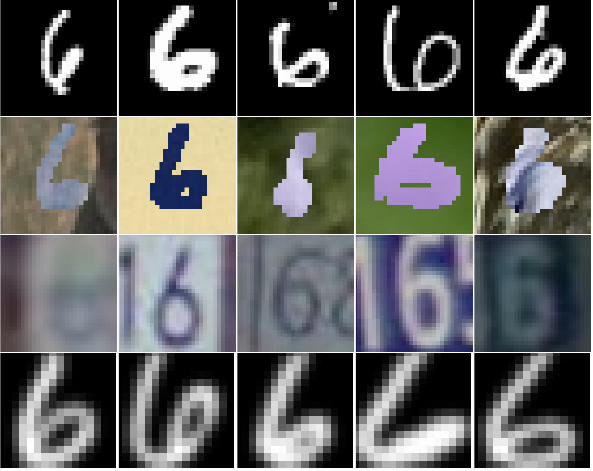}
    \caption{Twenty sample images from class \enquote{6} of the \combidigits dataset. Each row corresponds to one of the four source datasets: \mnist, \mnistm, \svhn, and \usps (top to bottom), illustrating the intra-class heterogeneity across domains.}
    \label{fig:combidigits-6}
\end{figure}

\paragraph{\combidigits (Ours).}
A synthetic dataset constructed by merging all classes from \mnist, \mnistm, \svhn, and \usps. Each class label corresponds to a digit (0–9) and \cref{fig:combidigits-6} shows example images of class \enquote{6}.
The dataset includes significant visual heterogeneity across sources and thus simulates multi-modal, non-collapsed class distributions.

\subsection{Imprinting as Memory}\label{ss:imprinting-as-memory}

We revisit the classical idea of Bidirectional Associative Memories (BAMs)~\citep{kohonen,anderson72,nakano,anderson77} and the associated update rule~\citep{sejnowski,dayan} for storing key-value pairs as in

\[
\mathbf{W} \leftarrow \mathbf{W} + \vect{v} \vect{k}^\top,
\]

where $\mathbf{W}\in\R^{l \times l}$ is a matrix and $\vect{k},\vect{v}\in\R^d$ are key and value vectors, respectively, to be stored therein.

\par
Throughout this work, imprinting can be interpreted as inserting such key-value associations into the classifier weights.
More precisely, the generation (\GEN) component extends the linear classification head by setting \(\vect{v}\) as a one-hot vector representing the class, and \(\vect{k}\) as the corresponding proxy previously computed.
While we focus on this simple linear setting, imprinting is not limited to it, as discussed in \cref{sec:related}.

\par
Notably, this form of direct memory update has seen renewed attention in modern architectures beyond standard query-key-value attention.
In particular, the xLSTM model~\citep{xlstm} implements this mechanism within its mLSTM memory blocks, where the matrix memory cell is updated by gated key-value associations, closely following this classical covariance rule, indicating a broader resurgence of associative memory principles in contemporary sequence modeling.

\subsection{Differences between Foundation Models}\label{ss:archidifferences}
While an in-depth comparison of foundation models is beyond the scope of this work, we believe it is important to highlight key observations we have made.
In particular, \cref{fig:nc1s} shows significantly lower \ncone scores for \vitb and \swinb on their pre-training \imagenet data compared to the \texttt{resnet} models.
We hypothesize that this difference is primarily due to model size and training regimes.
The Transformer-based architectures (\vitb and \swinb) have a considerably higher parameter count ($\approx 87$M) than the \texttt{resnet} models (11.7M and 25.6M, respectively). Additionally, \vitb and \swinb were trained for more than three times as many epochs (300 vs. 90) while using a substantially lower learning rate (0.003 and 0.01 vs. 0.1).
Notably, the embedding dimensions of these models are comparable, meaning that the observed differences in \ncone scores cannot be attributed to differences in representation dimensionality.
Instead, we argue that the combination of larger model size, extended training duration, and lower learning rates likely contributes to greater overfitting, leading to more pronounced collapse.
As discussed in \cref{ss:nc-and-acc}, this enables the Transformer-based \model{s} to handle the \imagenet tasks with remapped labels more effectively and to achieve significantly better performance on the similarly distributed \cifar dataset.

\begin{figure}[tb]
    \centering
    \begin{subfigure}[t]{0.49\linewidth}
        \vspace{0pt}
        \centering
        \includegraphics[width=\linewidth]{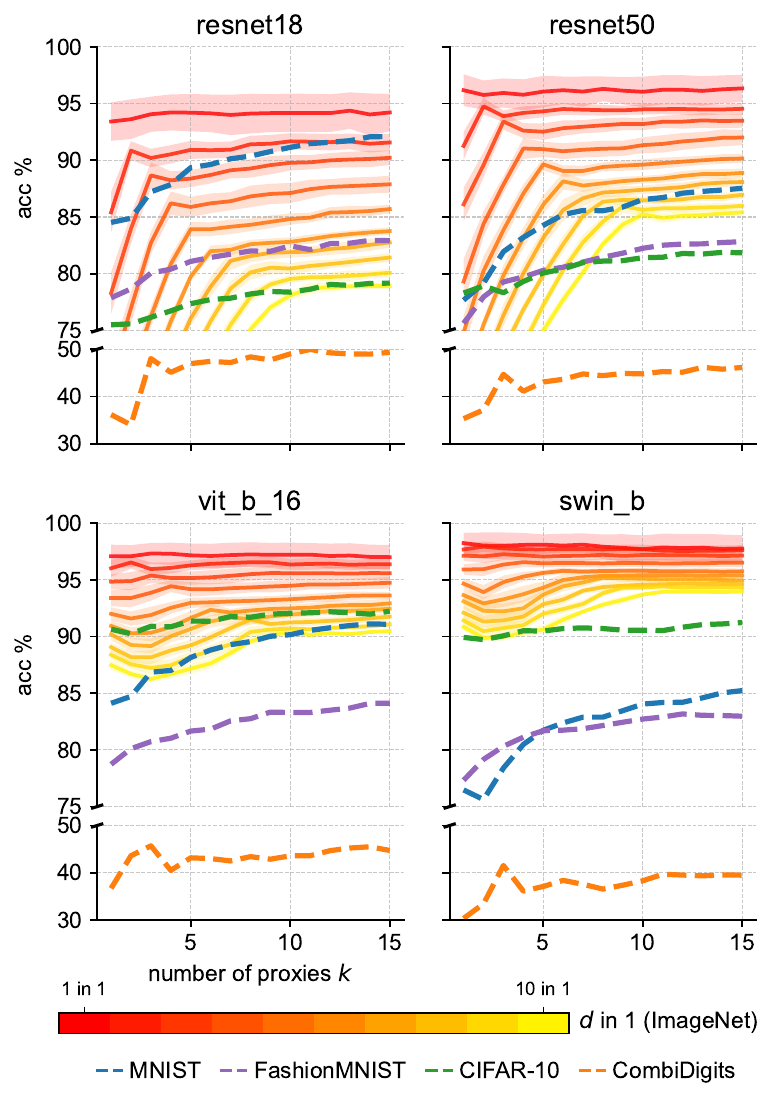}
        \caption{
        \kmeans used as \GEN.
        \NORMpost and \NORMinf are set to \ll, and \NORMpre to \none.
        Besides the prominent peaks in accuracy at $k=d$ (as already observed in \cref{fig:label-remappings-task-acc-vs-num-proxies-qinorm}), a consistent dip between $k=1$ and $k=d$ appears in Transformer-based models, which was not to be seen with \NORMpre set to \ll as well.
        In \cref{ss:archidifferences} we hypothesize that this is due to the distinct embeddings distributions of CNN- and Transformer-based architectures.
        }
        \label{fig:label-remappings-task-acc-vs-num-proxies-almost-qinorm}
    \end{subfigure}
    \hfill
    \begin{subfigure}[t]{0.49\linewidth}
        \vspace{0pt}
        \centering
        \includegraphics[width=\linewidth]{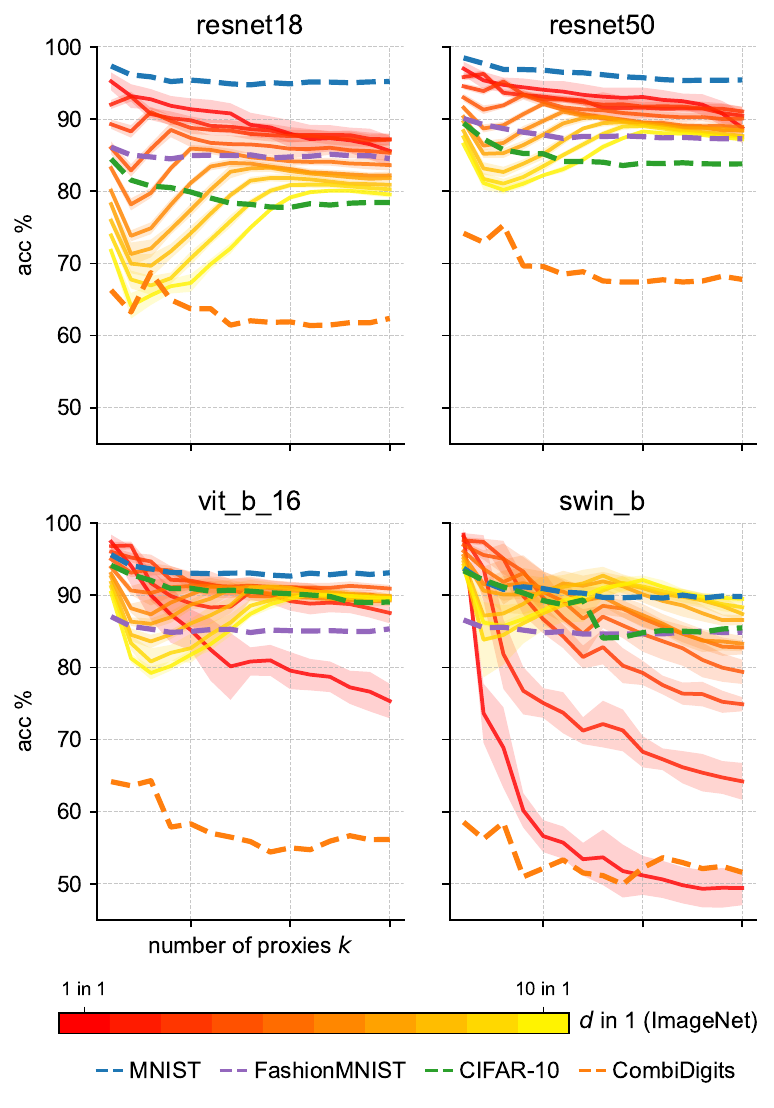}
        \caption{
        \kls (see \cref{ss:optimalweights}) used as \GEN and all \NORMs are set to \none.
        Here, the CNN-based \model{s} exhibit a clear accuracy peak at $k=d$, with a consistent drop for intermediate values $1<k<d$.
        For the Transformer-based \model{s}, however, accuracy is typically highest at $k=1$ and only improves slightly around $k=d$ but does not show a pronounced peak.
        For all \model{s}, performance declines with $k>d$, indicating that excessive proxy splitting harms generalization.
        This degradation does not occur in \kmeans \GEN, which remains robust even with large $k$ (see \cref{fig:label-remappings-task-acc-vs-num-proxies-qinorm,fig:label-remappings-task-acc-vs-num-proxies-almost-qinorm}).
        }
        \label{fig:label-remappings-task-acc-vs-num-proxies-kls}
    \end{subfigure}
    \caption{
        Averaged accuracy of ten random \imagenet label remappings (\enquote{$d$ in $1$}) for every $d = 1,\dotsc,10$ over number of proxies $k$.
        Shaded areas indicate $95\%$ CIs over three different seeds.
        Values for the tasks containing all of \mnist, \fmnist, \cifar, resp. \combidigits at once are shown in dotted lines.
    }   
\end{figure}

\Cref{fig:label-remappings-task-acc-vs-num-proxies-almost-qinorm}, similar to \cref{fig:label-remappings-task-acc-vs-num-proxies-qinorm}, illustrates the impact of varying the number of proxies on imprinting accuracy across different foundation models (\model{s}).
The key difference in this figure is the use of \none for \NORMpre instead of \ll.
This seemingly minor change reveals a striking contrast between CNN- and Transformer-based architectures: a distinct and consistent dip between $k=1$ and $k=d$ appears in Transformer-based models, whereas this dip is absent in \cref{fig:label-remappings-task-acc-vs-num-proxies-qinorm}, where \ll is used as \NORMpre, and does not occur at all in the \texttt{resnet} models.
We hypothesize that this difference arises from the distinct embedding distributions of CNN- and Transformer-based architectures (see, e.g., \citet[Figure S2]{itsdone}).

\subsection{Learned Weights (with Multiple Proxies) and Comparisons}\label{ss:optimalweights}

\citet{harun2025good} recently studied the initialization of classifier weights for novel categories in a Continual Learning (CL) setting using the least squares algorithm, comparing it to random initialization and class mean imprinting across various loss functions.
In contrast, our work centers on imprinting-based approaches, which avoid using gradient-based optimization and cross-class statistics, operating instead on a per-class basis with immediate availability of data.
Nonetheless, we include the least-squares method as a supervised oracle baseline -- explicitly not an imprinting method -- fine-tuned for classification accuracy.

We evaluate the performance of \ls on the same tasks as defined in \cref{sec:experiments}.
Additionally, we introduce a multi-proxy extension, \kls, which combines least squares with \kmeans clustering.
As in the case of weight generation \GEN in imprinting (where \kmeans outperforms \mean), we find that using multiple proxies per class improves performance in settings with high \ncone as well, indicating that the benefits of multi-prototype representations persist even in this non-imprinting, supervised context.

All experiments use no normalization (\none as \NORMs), based on ablations  confirming that additional normalization impairs performance.
This matches expectations, since least squares outputs are already calibrated, and normalization distorts them.

\paragraph{Least Squares Weights.}
For all data contained in task \task, define $\mathbf{H} \in \mathbb{R}^{l \times N}$ as the collection of all feature vectors of the $N$ training samples in \task obtained by applying a fixed \model{}, i.e., $\vect{h}_{c,i} \in \mathbb{R}^{l}$ is the feature vector of the $i$-th sample in the $c$-th class.
Recall from \cref{ss:quantnc} that the within-class covariance matrix $\mathbf{\Sigma}_W$ is given as $\mathbb{E}_{c,i}[(\vect{h}_{c,i} - \overline{\mathbf{h}}_c)(\vect{h}_{c,i} - \overline{\mathbf{h}}_c)^\top]$ and additionally define the total covariance matrix $\mathbf{\Sigma}_T$ and the class-means matrix $\mathbf{M}$ as
\[
\mathbf{M} = [\overline{\vect{h}}_1, \ldots, \overline{\vect{h}}_C] \in \mathbb{R}^{l \times C}, \qquad \mathbf{\Sigma}_T = \mathbb{E}_{c,i}[(\vect{h}_{c,i} - \vect{h}_G)(\vect{h}_{c,i} - \vect{h}_G)^\top] \in \mathbb{R}^{l \times l}.
\]

From these feature statistics, we can obtain the \ls weights via
\begin{equation}
\label{eq:w_ls}
    \mathbf{W}_{LS} = \frac{1}{C} \mathbf{M}^\top (\mathbf{\Sigma}_T + \vect{h}_G \vect{h}_G^\top + \lambda \mathbf{I})^{-1}.
\end{equation}

Here, $\mathbf{I}$ is the identity matrix and $\lambda$ is the weight decay.
We set $\lambda$ to match the value used during the original training of each respective model.
\Cref{tab:weight-decay} lists the specific $\lambda$ values used for all models considered in our experiments.

\begin{table}[htbp]
    \captionsetup{width=\textwidth}
    \caption{Weight decay values $\lambda$ used for each model.}
    \centering
    \begin{tabular}{lc}
        Model & \(\lambda\) \\
        \midrule
        \resnetsmall, \resnetlarge~\citep{resnet} & 0.0001 \\
        \vitb~\citep{vit} & 0.1 \\
        \swinb\citep{swin} & 0.05 \\
    \end{tabular}
    \label{tab:weight-decay}
\end{table}

\Cref{eq:w_ls} reveals that \ls shares structural similarities with \mean imprinting, but applies a more sophisticated normalization scheme that includes both scaling and rotation.
Crucially, it relies on cross-class statistics computed across the entire dataset, and thus is incompatible with imprinting scenarios, which operate by directly constructing class weights from the data of a single class without supervision or access to other classes.

\begin{algorithm}[tbph]
\caption{\kls}
\label{alg:kls}
\begin{algorithmic}[1]
\Require Class data $\{\mathbf{H}_c\}_{c=1}^C$, number of proxies $k$
\Ensure Weights $\{\mathbf{W}_c\}_{c=1}^C$ with shape $[k, l]$ per class
\If{$k = 1$}
    \State \Return standard least squares weights $\mathbf{W}_{LS}$ for each class (see \cref{eq:w_ls})
\EndIf
\For{each class $c$}
    \State Cluster $\mathbf{H}_c$ into $k$ clusters via $k$-means
    \State Assign each cluster $j$ to proxy class $(c, j)$
    \State Collect proxy samples $\{\mathbf{H}_{(c,j)}\}_{j=1}^{k}$
\EndFor
\State Compute least squares weights $\vect{w}_{(c,j)}$ jointly for all proxy classes $(c,j)$
\State Assemble $\mathbf{W}_c = [\vect{w}_{(c,1)}, \ldots, \vect{w}_{(c,k)}]$ for each original class $c$
\State \Return $\{\mathbf{W}_c\}_{c=1}^C$
\end{algorithmic}
\end{algorithm}

\paragraph{Combining \kmeans and \ls into \kls.}

To allow for multiple proxies per class, we propose \kls, a generalization of standard least squares (\ls) that integrates clustering.
Instead of computing a single weight vector per class using all class samples, we first partition each class’s feature set $\mathbf{H}_c$ into $k$ clusters using $k$-means.
Each cluster is treated as a separate proxy class, effectively expanding the classification task from $C$ to $k \cdot C$ proxy classes.
We then solve a single regularized least squares problem over all proxy classes at once, assigning a distinct target vector to each proxy.
The resulting weights are grouped by their original class to yield $k$ weight vectors per class.
The complete procedure is given in \cref{alg:kls}.

\par
The results in \cref{fig:label-remappings-task-acc-vs-num-proxies-kls} extend previous findings from \mean and \kmeans for the CNN-based \model{s}, as accuracy increases with increasing proxy count $k$, peaks at $k = d$, and reveals a distinct dip between $k=1$ and $k=d$, confirming the weakness of using only one proxy per class.
For Transformer-based \model{s}, by contrast, accuracy is often highest at $k=1$, improves slightly near $k=d$, but lacks a pronounced peak.
Beyond $k>d$, accuracy generally declines for \kls, indicating diminishing returns from excessive proxy splitting.
In contrast, \kmeans \GEN maintains stable performance even as $k$ grows, demonstrating robustness to increasing proxy counts.

\paragraph{Comparison with \kmeans.}

We now compare the performance of \kmeans and \kls as proxy generation methods, noting again that \kls is not an imprinting scheme as it does not operate on a class-by-class basis as the imprinting methods presented in \cref{sec:method} do.

\begin{figure}[tbhp]
    \centering
    \includegraphics[width=0.8\linewidth]{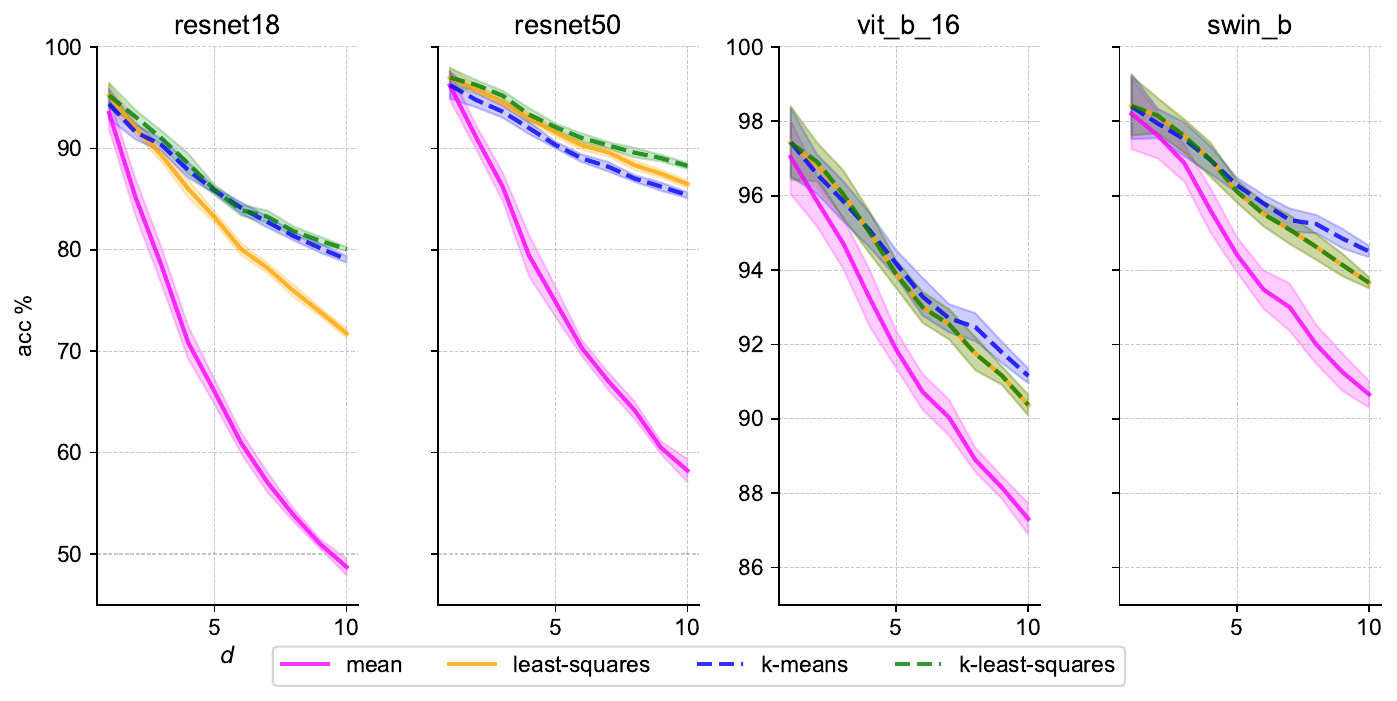}
    \caption{Averaged accuracy of ten random \imagenet label remappings (\enquote{$d$ in $1$}) for every $d = 1,\dotsc,10$ for \mean and \ls, and optimal \kmeans and \kls ($k\in\{0,\dotsc,15\}$) weights.
    Shaded areas indicate $95\%$ CIs over three different seeds.
    Note the differing y-axis scales for CNN-based (\resnetsmall, \resnetlarge) and Transformer-based (\vitb, \swinb) models.
    The figure illustrates that with increasing multi-modality $d$, multi-proxy methods (\kmeans, \kls) generally outperform single-proxy methods (\mean, \ls), and that the \kmeans imprinting scheme is competitive with \ls approaches.
    }
    \label{fig:kls_comp}
\end{figure}

Upon proper comparison of the best performing \kmeans and \kls configurations, as depicted in \cref{fig:kls_comp}, we observe that on our synthetic \imagenet tasks, \kls does not consistently offer a substantial improvement.
In fact, it is only on \resnetlarge, in a highly multi-modal setting ($d=10$), that \kls reaches better accuracy.
For the Transformer-based models, \kmeans even performs better in these scenarios.

While the Transformer-based models generally achieve superior performance overall (note the different y-axis scales for CNN- and Transformer-based models in \cref{fig:kls_comp}), a consistent trend across all models is observed: with increasing multi-modality (increasing $d$), using more than one proxy, whether through \kmeans or \kls, begins to outperform the single-proxy methods (\mean and \ls).
Still, it is striking to see how effective single weights with \ls can be.
However, it is also noteworthy how easily \kmeans can bridge this gap, particularly considering that \ls is not an actual imprinting scheme, while \kmeans is.

Furthermore, we evaluate the \kls weight generation method (using \none for all \NORMs) across the 48 tasks proposed in the main part of our paper (see \cref{sec:experiments}).
\ls reaches an average accuracy of $94.54$\%, while \texttt{20-least-squares} \textbf{drops} to $91.20$\%.
The observed decrease in accuracy can be explained by the dynamics shown in \cref{fig:label-remappings-task-acc-vs-num-proxies-kls}.
In comparison to the numbers presented in \cref{tab:special-cases}, this shows that our $k$-means imprinting scheme significantly bridges the gap between single-proxy imprinting ($k=1$) and the optimal least squares weights.

\subsection{Computational Efficiency}\label{ss:comp_efficiency}

\paragraph{Clustering-based imprinting.}
Our \kmeans implementation in \GEN uses \texttt{sklearn.cluster.KMeans} with its default parameters.
The cost is $\mathcal{O}(Nklt)$ for $N$ samples, $k$ clusters, feature dimension $l$, and $t$ Lloyd iterations.
While $k$-means is not computationally negligible, both assignment and update steps in each iteration parallelize naturally, and convergence typically requires only a few iterations.
Similarly, the other steps in our imprinting framework (\NORMs, \AGG) scale linearly with dataset size or proxy count and parallelize efficiently, resulting in no practical scalability bottlenecks even for larger or more complex datasets.

\paragraph{Gradient-based optimization.}
The closed-form \ls costs $\mathcal{O}(Nl^2 + l^3 + NlC)$ from the covariance computation and matrix inversion.
If solved through stochastic gradient descent, the cost is $\mathcal{O}(NlCt)$ with $t$ epochs.

\paragraph{Empirical runtime.}
\Cref{tab:gen-runtime} shows that \ls is faster than \kmeans in practice, likely because it processes all data in a single closed-form step rather than iterating class-by-class.
Nevertheless, it should be clarified again that \ls represents the analytic, non-iterative optimum derived from gradient minimization and lacks sequential class handling -- precisely where imprinting excels (e.g., in continual or edge-learning scenarios).

\begin{table}[htbp]
    \captionsetup{width=\textwidth}
    \caption{
        Average runtime in seconds for different \GEN variants across all 48 tasks (see \cref{sec:experiments}).
        To ensure transparency, all timing measurements were obtained on identical hardware (8 vCPUs on Intel Xeon Gold 6438Y+ nodes (2 sockets, 64 physical cores/128 threads)).
        Cores were not pinned to processes, which may introduce minor variance; however, we observed less than 10\% variation across seeds.
    }
    \centering
    \begin{tabular}{lc}
        \toprule
        \GEN variant & Runtime (s) \\
        \midrule
        \mean & 0.0029 \\
        \texttt{5-means} & 1.2249 \\
        \texttt{20-means} & 1.7321 \\
        \ls & 0.1593 \\
        \texttt{5-least-squares} & 1.3724 \\
        \texttt{20-least-squares} & 2.3717 \\
        \bottomrule
    \end{tabular}
    \label{tab:gen-runtime}
\end{table}


\end{document}